\documentclass[twocolumn, switch]{article}
\usepackage{preprint}
\usepackage{amsmath, amsthm, amssymb, amsfonts}
\usepackage{bm}
\usepackage{mathtools}
\usepackage{algorithmic}
\usepackage{algorithm}
\usepackage{array}
\usepackage[caption=false,font=normalsize,labelfont=sf,textfont=sf]{subfig}
\usepackage{textcomp}
\usepackage{stfloats}
\usepackage{url}
\usepackage{verbatim}
\usepackage{graphicx}
\usepackage{cite}
\usepackage{color}
\usepackage{multirow}
\usepackage{hhline}
\usepackage{graphicx}
\usepackage{tabularray}
\usepackage{threeparttable}

\usepackage[numbers,square]{natbib}
\usepackage[utf8]{inputenc}	
\usepackage[T1]{fontenc}	
\usepackage{xcolor}		
\usepackage[colorlinks = true,
            linkcolor = blue,
            urlcolor  = blue,
            citecolor = blue,
            anchorcolor = black]{hyperref}	
\usepackage{booktabs} 		
\usepackage{nicefrac}		
\usepackage{microtype}		
\usepackage{lineno}		
\usepackage{float}			

\usepackage{titlesec}
\titlespacing\section{0pt}{12pt plus 3pt minus 3pt}{1pt plus 1pt minus 1pt}
\titlespacing\subsection{0pt}{10pt plus 3pt minus 3pt}{1pt plus 1pt minus 1pt}
\titlespacing\subsubsection{0pt}{8pt plus 3pt minus 3pt}{1pt plus 1pt minus 1pt}

\newcommand{\etal}{\textit{et al}.~}
\newcommand{\eq}[1]{Eq.~(#1)}
\newcommand{\ie}{\textit{i}.\textit{e}.}
\newcommand{\eg}{\textit{e}.\textit{g}.}
\newcommand{\etc}{\textit{etc}.}

\newcommand{\argmin}[1]{\underset{#1}{\operatorname{arg}\,\operatorname{min}}\;}

\newcommand{\fig}[1]{Fig.~(#1)}

\title{Learning Many-to-Many Mapping for Unpaired Real-World Image Super-resolution and Downscaling}

\usepackage{authblk}

\author{Wanjie Sun}
\author{Zhenzhong Chen\thanks{\tt{zzchen@whu.edu.cn}}}
\affil{School of Remote Sensing and Information Engineering, Wuhan University}

\begin{document}
\twocolumn[ 
  \begin{@twocolumnfalse} 

\maketitle

\begin{abstract}
Learning based single image super-resolution (SISR) for real-world images has been an active research topic yet a challenging task, due to the lack of paired low-resolution (LR) and high-resolution (HR) training images. Most of the existing unsupervised real-world SISR methods adopt a two-stage training strategy by synthesizing realistic LR images from their HR counterparts first, then training the super-resolution (SR) models in a supervised manner. However, the training of image degradation and SR models in this strategy are separate, ignoring the inherent mutual dependency between downscaling and its inverse upscaling process. Additionally, the ill-posed nature of image degradation is not fully considered. In this paper, we propose an image downscaling and SR model dubbed as SDFlow, which simultaneously learns a bidirectional many-to-many mapping between real-world LR and HR images unsupervisedly. The main idea of SDFlow is to decouple image content and degradation information in the latent space, where content information distribution of LR and HR images is matched in a common latent space. Degradation information of the LR images and the high-frequency information of the HR images are fitted to an easy-to-sample conditional distribution. Experimental results on real-world image SR datasets indicate that SDFlow can generate diverse realistic LR and SR images both quantitatively and qualitatively.
\end{abstract}

  \end{@twocolumnfalse} 
] 

\section{Introduction}

Image super-resolution (SR) is a fundamental and active research topic with a long-standing history in low-level computer vision \cite{yang2017image}, where single image super-resolution (SISR) received more attention for its wide applications in practical scenarios. SR aims to reconstruct or estimate the realistic underlying HR images from the given LR images, either for perceptual viewing purposes or to help improve other downstream vision tasks. SISR is an ill-posed problem where multiple HR images can be matched to the same LR image. Benefiting from the recent advancements of deep convolutional neural network (CNN) techniques, the performance of CNN-based SR models has significantly surpassed that of the interpolation-based algorithms \cite{yang2019deep,wang2020deep,chen2022real,liu2022blind}. For many years, most existing CNN-based SR models are trained on artificially synthesized paired low-resolution (LR) and high-resolution (HR) image datasets, leading to poor performance in super-resolving the real-world LR images due to the domain gap between the synthesized training data and the real-world test data. Thus, more attention has been drawn to developing SR approaches for real-world images \cite{chen2022real}, promoting CNN-based SR in practical applications.

Real-world single image super-resolution (RSISR) is a challenging task, due to the lack of paired real-world LR/HR training images. The most straightforward solution is to capture real-world LR/HR image pairs by changing the focal length of a zoom lens camera to shoot the same scene \cite{wei2020component,chen2019camera,cai2019toward,city100}. Some efforts are taken to capture more accurately paired real-world LR/HR image pairs using specially designed devices, such as hardware binning \cite{kohler2019toward} and beam splitter \cite{joze2020imagepairs}. However, these approaches are laborious and costly, meanwhile, the degradation types are bound by the acquisition devices. Additionally, perfectly aligned real-world LR/HR image pairs are hard to construct due to the imperfection and characteristics of optical imaging device \cite{sun2021learning}, hampering the performance of pixel-wise loss function widely used in SR model training. As an alternative, most of the RSISR works focus on learning from unpaired real-world LR/HR images that are easily accessible \cite{chen2022real}. This kind of work usually adopts a two-stage training strategy \cite{son2021toward,sun2021learning,wei2021unsupervised,maeda2020unpaired,lugmayr2019unsupervised,fritsche2019frequency,yuan2018unsupervised,bulat2018learn}, where a downscaling network is first trained under the generative adversarial network (GAN) \cite{goodfellow2014generative} framework to approximate the distribution of real-world LR images given HR images. Then SR models are trained in a supervised manner on the synthetic real-world LR/HR image pairs constructed using the trained downscaling network. However, the two-stage strategy has common shortcomings \cite{romero2022unpaired}. First, the downscaling network is usually optimized for the deterministic generation of LR images from HR inputs without considering that real-world image degradations are stochastic, \ie, an HR image can be degraded into multiple LR images, which may limit the performance of the following SR models. Second, although some recent works \cite{wolf2021deflow,ning2022learning,luo2022learning,lee2022learning} try to learning to generate diverse degraded LR images from its HR counterpart, however the training of the downscaling network and SR network is separate. Since the image downscaling and SR are inherently an inverse problem, thus the mutual dependency between image downscaling and SR is not fully exploited in the two-stage methods.

\begin{figure}
    \centering
    \includegraphics[width=\columnwidth]{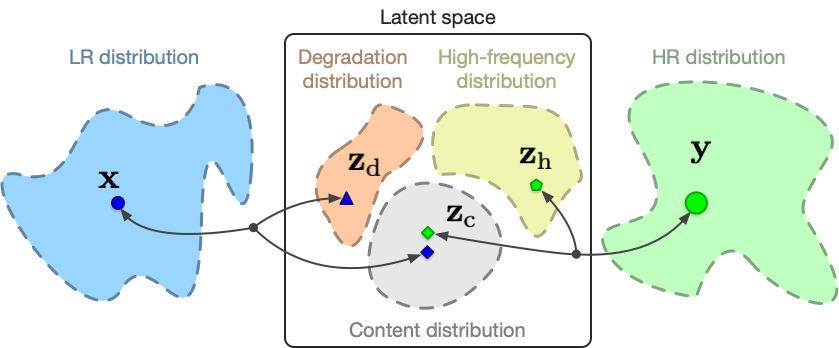}
    \caption{The graphical model of the SDFlow. It simultaneously learns the diverse generation of real-world LR and SR images as a unified task.}
    \label{fig:main_idea}
\end{figure}

In this paper, we address the aforementioned problem by proposing a novel invertible model, named as SDFlow, simultaneously learning diverse image SR and diverse image downscaling (degradation) under the normalizing flow \cite{Laurent2015,Laurent2017,kingma2018glow} framework. As shown in \fig{\ref{fig:main_idea}}, the main idea of the SDFlow is to discover the shared latent structure of real-world HR image and LR images by decoupling image content and its domain-specific variations in the latent space. Specifically, we first introduce two different latent variables for LR images which encode their contents and degradations, respectively. For HR images, we assume that they can be decoupled into content representation and residual high-frequency (HF) component. Content latent variables are learned to have shared representation so that the unpaired real-world LR images and HR images can be mapped to a domain invariant space. Degradation variations and HF components are mapped to an easy-to-sample latent distribution through a series of invertible transformations. Considering the complexity of real-world degradations, we model the degradation distribution using a mixture of Gaussian distributions. During inference, the content latent variable is first obtained, then diverse LR or SR images can be reconstructed by sampling the degradation or HF latent space, respectively. Contrary to the two-stage method, we reformulate the real-world SISR as a dual modeling problem of both real-world image downscaling and SR as a unified task. As the downscaling process and the SR process share the same model parameter, the SDFlow model can be optimized jointly in both directions, forming an implicit constraint on the solution space of both the image downscaling and SR \cite{guo2020closed}. To distinguish from the image rescaling model based on the invertible neural network (INN) \cite{xiao2020invertible,liang2021hierarchical}, which can only well reconstruct the LR image downscaled by the forward process. Our proposed SDFlow can generate multiple realistic LR images, while it can also estimate diverse HR images with better perceptual quality from the real-world LR input. To the best of our knowledge, our work is the first attempt to simultaneously model diverse image SR and downscaling from unpaired real-world image data in a unified framework.

The rest of the paper is organized as follows: Section II reviews the work of RSISR and normalizing flow for computer vision. Section III first formulates the problem of many-to-many mapping between real-world image SR and image downscaling, then introduces the network structure of the proposed SDFlow. Section IV evaluates and analyzes the experimental results of the SR and LR images quantitatively and qualitatively. Finally, Section V summarizes our work.

\section{Related Work}
\subsection{Real-world single image super-resolution}
One of the challenges hindering the applications of most state-of-the-art CNN-based SISR models in practical scenarios is the lack of paired real-world LR/HR training data. Thus, some efforts are devoted to collecting paired real-world LR/HR images by capturing the same scene with different resolutions using zoom lens camera \cite{cai2019toward,wei2020component,chen2019camera} or specific imaging device \cite{kohler2019toward,joze2020imagepairs}. For example, the RealSR dataset \cite{cai2019toward}, the DRealSR dataset \cite{wei2020component} and the City100 dataset \cite{chen2019camera} are constructed by adjusting the focal length or shooting distance to obtain LR/HR image pairs of the same scene. While the SuperER dataset \cite{kohler2019toward} and the ImagePairs \cite{joze2020imagepairs} dataset are constructed using hardware binning and optical beam splitter, respectively. However, collecting the paired data is time-consuming and effort-consuming, and it often requires a strictly controlled environment. Besides, miss-aligned LR/HR pairs are unavoidable due to the imperfection of imaging system, such as lens distortion, and the depth-of-field phenomenon.

To circumvent the above mentioned issues, several RSISR works propose to synthesize real-world LR images from their HR counterparts. One branch of approaches \cite{ji2020real,zhou2019kernel} assumes that real-world degradations can be modeled as blur kernels and additive noises. Zhou \etal \cite{zhou2019kernel} first build a large kernel pool via kernel estimation from real photographs, then SR networks are trained using the LR/HR pairs constructed using the generated kernels. Ji \etal \cite{ji2020real} further estimates the real noise distribution to better imitate real-world LR images. Another branch of methods \cite{bulat2018learn,yuan2018unsupervised,lugmayr2019unsupervised,fritsche2019frequency,sun2021learning,wei2021unsupervised,maeda2020unpaired} proposes to use GANs to unsupervisedly learn the conditional distribution of LR images given HR inputs. Yuan \etal \cite{yuan2018unsupervised} propose CinCGAN to learn the distribution of clean LR and HR image separately, which guides the SR of unpaired LR images. Bulat \etal \cite{bulat2018learn} first trains a high-to-low GAN with unpaired HR and LR images, then the output of the network, acting as the paired LR image, is used to train the SR network. Lugmayr \etal \cite{lugmayr2019unsupervised} and Maeda \etal \cite{maeda2020unpaired} use CycleGAN to translate HR images to LR images and consequently constructed the paired LR and HR dataset that is used for training of the SR network. In \cite{fritsche2019frequency}, Fritsche \etal proposed to separate the low and high frequencies, where the HF information is adversarially trained, and the low-frequency information is learned with the L1 loss. Instead of assuming that domain gap between Bicubic downscaled (BD) LR images and real LR images can be eliminated by GANs, Wei \etal \cite{wei2021unsupervised} and Sun \etal \cite{sun2021learning} further adopt a domain distance aware training strategy by incorporating domain distance information in the training process of SR networks.

Existing RSISR methods trained on sets of paired or unpaired LR/HR images suffer from inconsistency problems between test and training data. To reduce the impact on SR performance due to inconsistency between training and test data distribution, the self-similarity property of the natural image is exploited to learn image-specific SR models. Shocher \etal \cite{shocher2018zero} proposed the ZSSR model utilizing cross-scale internal patch recurrence of the natural image to construct test image-specific LR/HR pair for SR training. Besides, data augmentation is adopted to enrich training pairs. Bell-Kligler \etal \cite{bell2019blind} proposed to train an image-specific GAN (KernelGAN) to model the blur kernel of the LR input to further generate more realistic degraded LR/HR pairs. Thus, fully self-supervised image-specific RSISR is achieved by plugin the KernelGAN into the ZSSR. Kim \etal \cite{kim2020dual} developed the DBPI, a unified internal learning-based SR framework consisting of an SR network and a downscaling network to jointly train the image-specific degradation and SR networks. The SR network is trained to recover the LR input from its downsampled version generated by the downscaling network. meanwhile, the downscaling network is optimized to restore the LR input from its HR version reconstructed by the SR network. Similarly, DualSR proposed by Emad \etal \cite{emad2021dualsr} jointly optimizes the image-specific downsampler and the upsampler. By using the patches from the test image, the DualSR is trained with cycle consistency, masked interpolation and adversarial loss. Recently, inspired by the Maximum A Posteriori estimation, Chen \etal \cite{chen2023self} proposed a self-supervised cycle-consistency based scale-arbitrary RSISR model.

\subsection{Normalizing flow and INN for low-level computer vision}
Normalizing flows are a family of generative model, where both sampling and density evaluation can be efficient and exact \cite{kobyzev2020normalizing}. Normalizing flows are usually constructed by a series of invertible transformations parameterized by neural networks. Normalizing flow transforms complex distribution into a simple tractable one in the forward process and achieves diverse generations in the reverse process by sampling from the simple distribution. Thus, normalizing flow is suitable for modeling the solution space of many ill-posed inverse problems \cite{ardizzone2018analyzing}. Recently, normalizing flow has demonstrated promising results for many ill-posed inverse tasks in low-level computer vision, such as image SR, denoising, colorization, low-light image enhancement and \etc. SRFlow \cite{lugmayr2020srflow} maps the distribution of HR images into an isotropic Gaussian distribution conditioned by the LR image in the forward process and achieves diverse SR reconstructions in the inverse process. HCFlow \cite{liang2021hierarchical} learns a bijective mapping between HR and LR image pairs by modeling the distribution of LR images close to that of the BD images, and the rest of HF distribution as Gaussian. FKP proposed by Liang \etal \cite{liang2021flow} improves the blind SR performance by modeling the complex kernel prior distribution using normalizing flow. Abdelhamed \etal \cite{abdelhamed2019noise} proposed a normalizing flow based noise modeling method that can produce realistic noise conditioned on the raw image and camera parameters. DeFlow \cite{wolf2021deflow} unsupervisedly learns stochastic image degradation in latent space using a shared flow encoder-decoder network. Yu \etal \cite{yu2020wavelet} proposed to transform the HF components of wavelet transformation into Gaussian distortion, achieving fast training and HR image generation using the multi-scale nature of wavelet transformation.

As the building blocks of the normalizing flow, INNs have recently been explored for other image processing tasks \cite{xiao2020invertible,xing2021invertible,liu2021invertible,zhao2021invertible}. Xiao \etal \cite{xiao2020invertible} first proposed to use INN to jointly model the image downscaling and upscaling, significantly improving the performance of the image rescaling task. Xing \etal \cite{xing2021invertible} proposed an invertible ISP model to generate visual pleasing sRGB image from RAW input, and recover nearly the same quality RAW data at the same time. Liu \etal \cite{liu2021invertible} models the image denoising using INN. The noised image is decouple into LR clean image and residual latent representations containing noise in the forward process, while the clean HR image is obtained in the reverse process by replacing the latent representations with another one sampled from a prior distribution. Similar ideas can be applied to the image de-colorization and re-colorization task. Zhao \etal \cite{zhao2021invertible} proposed to decouple the color image into grayscale and color information, modeled as Gaussian distribution, using the forward process of INN, meanwhile, the color image can be reconstructed in the inverse process by sampling the color information in the Gaussian space.
	
\section{Method}
In this section, we first formulate the problem of unpaired many-to-many mapping between image downscaling and SR, then we introduce the proposed SDFlow which can generate diverse realistic LR and SR images under a unified invertible neural network framework.

\subsection{Problem statement}
SISR algorithms try to estimate the underlying HR image corresponding to the single input LR image. This is a typical example of inverse problems, to solve which the forward process is generally required. Specifically, the degradation (downscaling) model is required in order to formulate what sort of data is to be used. The general image degradation model can be formulated as:
\begin{equation}\label{eq:deg_model}
    \mathbf{I}^\text{LR} = Deg(\mathbf{I}^\text{HR};\theta,s)
\end{equation}
where $\theta$ is the model parameters describing the degradation information, $s$ represents the downscaling factor. Practically, the degradation model is stochastic and complex, containing more than one type of degradations, such as blur, noise, compression artifacts and \etc \cite{wang2021real,zhang2022closer}. As can be inferred from \eq{\ref{eq:deg_model}}, the degradation is ill-posed since the degraded image can correspond to several degradation parameters $\theta$s. Thus, the degradation process is actually a one-to-many mapping.

With the defined forward process, the SISR aims to explicitly or implicitly invert the degradation process to estimate the HR image. This is achieved by optimizing a cost function defined from the perspective of either algebra or statistics. The commonly used Least-Squares cost function which minimized the $L_2$ norm is defined as:
\begin{equation}\label{eq:ls_sr}
    \hat{\mathbf{I}}^\text{HR} = \argmin{\mathbf{I}^\text{HR},\theta}||\mathbf{I}^\text{LR}-Deg(\mathbf{I}^\text{HR};\theta,s)||^2_2
\end{equation}
solving \eq{\ref{eq:ls_sr}} is challenging as it requires to estimate the unknown HR image and degradation parameters simultaneously, making the SISR ill-posed \cite{yang2017image}. In many real scenarios, there exist infinite combinations of HR images and degradation parameters that can produce the same LR image, thus, the SISR is also a one-to-many mapping.

Most SISR approaches tend to be supervised learning-based \cite{yang2019deep,wang2020deep}, requiring a well-defined degradation model to synthesize the LR images from HR images, so that the SISR algorithm knows what HR image should look like when super-resolving a given LR image. However, formulating the degradation model that can simulate the complex and stochastic image degradation process in real scenarios is challenging. Thus, synthesizing real-world LR/HR image pairs for supervised RSISR is a nontrivial task. Consequently, designing deep learning methods with unpaired data becomes an active research topic in RSISR and deserves in-depth exploration.

\subsection{Deriving the solution for many-to-many image SR and downscaling under the variational inference framework}
The learning settings we consider are as follows. We are given unpaired real-world LR and HR image datasets $\mathcal{X}=\{\mathbf{x}_i\}_{i=1}^N$ and $\mathcal{Y}=\{\mathbf{y}_j\}_{j=1}^M$ i.i.d. sampled from some true but unknown marginal distributions $p(\mathbf{X})$ and $p(\mathbf{Y})$, respectively. Our goal is to estimate the unknown conditional density $p(\mathbf{y}|\mathbf{x})$ and $p(\mathbf{x}|\mathbf{y})$ simultaneously, \ie, learning a invertible mapping $SD(\cdot)$ that can super-resolve a new LR image $\mathbf{x} \sim p(\mathbf{X})$ satisfying $SD(\mathbf{x}) \sim p(\mathbf{Y})$, while downscaling a new HR image $\mathbf{y} \sim p(\mathbf{Y})$ such that $SD^{-1}(\mathbf{y}) \sim p(\mathbf{X})$. However, it is difficult to model $p(\mathbf{y}|\mathbf{x})$ and $p(\mathbf{x}|\mathbf{y})$ only given marginal distributions $p(\mathbf{X})$ and $p(\mathbf{Y})$, thus, we seek to model the joint distribution $p(\mathbf{X}, \mathbf{Y})$ over $\mathcal{X}$ and $\mathcal{Y}$ under the unpaired learning setting. In general, this is a highly ill-posed problem, as the independence of $\mathbf{x}$ and $\mathbf{y}$ does not hold in practice \cite{wolf2021deflow}. We thus impose conditional independence by assuming that $\{\mathbf{x}, \mathbf{y}\}$ corresponds to three latent variables $\{\mathbf{z}_\text{c}, \mathbf{z}_\text{d}, \mathbf{z}_\text{h}\}$, where $\mathbf{z}_\text{c}$ is the underlying content representation shared by the LR and HR image, $\mathbf{z}_\text{d}$ is the degradation information contained in the LR image and $\mathbf{z}_\text{h}$ is the lost high-frequency information during the downscaling of the HR image. Under these assumptions, the conditional joint density can be represented as:
\begin{equation}\label{eq:conditional_independence}
    p(\mathbf{x}, \mathbf{y} | \mathbf{z}_\text{c}, \mathbf{z}_\text{d}, \mathbf{z}_\text{h}) = p(\mathbf{x} | \mathbf{z}_\text{c}, \mathbf{z}_\text{d})p(\mathbf{y} | \mathbf{z}_\text{c}, \mathbf{z}_\text{h})
\end{equation}
thus, we can maximize the likelihood of the joint density function $p(\mathbf{x}, \mathbf{y})$ using a variational posterior $q(\mathbf{z}_\text{c}, \mathbf{z}_\text{d}, \mathbf{z}_\text{h} | \mathbf{x}, \mathbf{y})$ whose log-likelihood is formulated as:
\begin{equation}\label{eq:factorization}
    \begin{aligned}
        \log q(\mathbf{z}_\text{c}, \mathbf{z}_\text{d}, \mathbf{z}_\text{h} | \mathbf{x}, \mathbf{y}) =& \log q(\mathbf{z}_\text{c} | \mathbf{x}, \mathbf{y}) + \log q(\mathbf{z}_\text{d} | \mathbf{z}_\text{c}, \mathbf{x}) \\
    & + \log q(\mathbf{z}_\text{h} | \mathbf{z}_\text{c}, \mathbf{y})
    \end{aligned}
\end{equation}

Since directly maximizing the log-likelihood of $p(\mathbf{x}, \mathbf{y})$ is generally intractable, we seek to maximize the evidence low bound (ELBO) under the variational inference framework:
\begin{equation}\label{eq:elbo}
    \begin{aligned}
        \log p(\mathbf{x}, \mathbf{y}) \ge& \mathbb{E}_{q(\mathbf{z}_\text{c}, \mathbf{z}_\text{d}, \mathbf{z}_\text{h} | \mathbf{x}, \mathbf{y})} [\log p(\mathbf{x}, \mathbf{y} | \mathbf{z}_\text{c}, \mathbf{z}_\text{d}, \mathbf{z}_\text{h})] \\
        & - D_\text{KL}(q(\mathbf{z}_\text{c}, \mathbf{z}_\text{d}, \mathbf{z}_\text{h} | \mathbf{x}, \mathbf{y}) || p(\mathbf{z}_\text{c}, \mathbf{z}_\text{d}, \mathbf{z}_\text{h}))
    \end{aligned}
\end{equation}
the right hand side of the \eq{\ref{eq:elbo}} is the ELBO, where the expectation term is the reconstruction error, and the second term of the ELBO minimizes the Kullback–Leibler (KL) divergence between the variational posterior and the prior of the latent variables. The reconstruction term with the conditional independence assumption in \eq{\ref{eq:conditional_independence}} and the factorization of the variational posterior in \eq{\ref{eq:factorization}} can be rewritted as:
\begin{equation}\label{eq:rewritted_reconstruction}
    \begin{aligned}
        &\mathbb{E}_{q(\mathbf{z}_\text{c} | \mathbf{x}, \mathbf{y}) q(\mathbf{z}_\text{d} | \mathbf{z}_\text{c}, \mathbf{x})}\log p(\mathbf{x} | \mathbf{z}_\text{c}, \mathbf{z}_\text{d}) \\
        &+ \mathbb{E}_{q(\mathbf{z}_\text{c} | \mathbf{x}, \mathbf{y}) q(\mathbf{z}_\text{h} | \mathbf{z}_\text{c}, \mathbf{y})}\log p(\mathbf{y} | \mathbf{z}_\text{c}, \mathbf{z}_\text{h})
    \end{aligned}
\end{equation}
next, we simplify the KL divergence term in \eq{\ref{eq:elbo}}. According to the graphical model presented in \fig{\ref{fig:main_idea}} and the conditional independence assumption, the prior $p(\mathbf{z}_\text{c}, \mathbf{z}_\text{d}, \mathbf{z}_\text{h})$ can be factorized into $p(\mathbf{z}_\text{c})p(\mathbf{z}_\text{d} | \mathbf{z}_\text{c})p(\mathbf{z}_\text{h} | \mathbf{z}_\text{c})$. Combining the factorized prior and \eq{\ref{eq:factorization}}, the KL divergence term can be re-formulated as:
\begin{equation}\label{eq:rewritted_kl}
    \begin{aligned}
        &D_\text{KL}(q(\mathbf{z}_\text{c} | \mathbf{x}, \mathbf{y}) || p(\mathbf{z}_\text{c})) + D_\text{KL}(q(\mathbf{z}_\text{d} | \mathbf{z}_\text{c}, \mathbf{x}) || p(\mathbf{z}_\text{d} | \mathbf{z}_\text{c})) \\
        &+ D_\text{KL}(q(\mathbf{z}_\text{h} | \mathbf{z}_\text{c}, \mathbf{y}) || p(\mathbf{z}_\text{h} | \mathbf{z}_\text{c}))
    \end{aligned}
\end{equation}
it is worth mentioning that the term $q(\mathbf{z}_\text{c} | \mathbf{x}, \mathbf{y})$ in \eq{\ref{eq:rewritted_reconstruction}} and \eq{\ref{eq:rewritted_kl}} still requires paired LR/HR images to infer the shared content latent variables. To meet the unpaired learning setting, we need to decouple the paired relationship, meaning that for the paired data $(\mathbf{x}, \mathbf{y})$, the content representation can be obtained from either LR image $\mathbf{x}$ or HR image $\mathbf{y}$ independently. Formally, this is written as:
\begin{equation}
    q(\mathbf{z}_\text{c} | \mathbf{x}, \mathbf{y}) = q(\mathbf{z}_\text{c} | \mathbf{x}) = q(\mathbf{z}_\text{c} | \mathbf{y}), \; \{\mathbf{x}, \mathbf{y}\} \sim p(\mathbf{x}, \mathbf{y})
\end{equation}
this requires that $\mathbf{z}_\text{c}$ encodes the domain invariant features between the LR and HR image domains. However, strictly ensuring this inference invariant condition is a difficult problem due to the unknown joint distribution $p(\mathbf{x}, \mathbf{y})$, we thus only require that the marginal distributions of $q(\mathbf{z}_\text{c} | \mathbf{x})$ and $q(\mathbf{z}_\text{c} | \mathbf{y})$ be aligned in the shared representation space, which can be achieved using the adversarial learning strategy described in Section \ref{sec:model_architecture}.

With the above derivation and simplification of \eq{\ref{eq:elbo}}, the expression of the final ELBO is:
\begin{equation}\label{eq:final_elbo}
\begin{gathered}
        \underbrace{\begin{aligned}
    &\mathbb{E}_{q(\mathbf{z}_\text{c} | \mathbf{x}) q(\mathbf{z}_\text{d} | \mathbf{z}_\text{c}, \mathbf{x})}\log p(\mathbf{x} | \mathbf{z}_\text{c}, \mathbf{z}_\text{d}) - D_\text{KL}(q(\mathbf{z}_\text{c} | \mathbf{x}) || p(\mathbf{z}_\text{c}))\\
        &- D_\text{KL}(q(\mathbf{z}_\text{d} | \mathbf{z}_\text{c}, \mathbf{x}) || p(\mathbf{z}_\text{d} | \mathbf{z}_\text{c}))
\end{aligned}}_{\text{ELBO}_\mathbf{x}}\\
    +\\
    \overbrace{\begin{aligned}
        &\mathbb{E}_{q(\mathbf{z}_\text{c} | \mathbf{y}) q(\mathbf{z}_\text{h} | \mathbf{z}_\text{c}, \mathbf{y})}\log p(\mathbf{y} | \mathbf{z}_\text{c}, \mathbf{z}_\text{h}) - D_\text{KL}(q(\mathbf{z}_\text{c} | \mathbf{y}) || p(\mathbf{z}_\text{c}))\\
        &- D_\text{KL}(q(\mathbf{z}_\text{h} | \mathbf{z}_\text{c}, \mathbf{y}) || p(\mathbf{z}_\text{h} | \mathbf{z}_\text{c}))
    \end{aligned}}^{\text{ELBO}_\mathbf{y}}
\end{gathered}
\end{equation}
\eq{\ref{eq:final_elbo}} can be grouped into two parts, the $\text{ELBO}_\mathbf{x}$ part only relates to the LR image while the $\text{ELBO}_\mathbf{y}$ part only relates to the HR image. This effectively removes paired training data requirements in maximizing the ELBO. In the next section, we introduce the SDFlow to learn the joint density function using \eq{\ref{eq:elbo}} and \eq{\ref{eq:final_elbo}}, which achieves many-to-many mapping between real-world image SR and downscaling.

\subsection{SDFlow}
Image downscaling and SR are inherently reciprocal with bi-directional ill-posedness property in real-world scenarios. The natural approach to tackle the ill-posed inverse problem is to consider the solution as a distribution that describes all of the possible results. We thus propose to simultaneously model the bidirectional one-to-many mapping between image downscaling and SR using the INN under the probabilistic normalizing flow framework, named SDFlow. When compared to other generative models such as GANs and VAEs in image downscaling and SR, flow-based models offer several advantages. As the model parameters of the INN are shared for both forward and backward process, the solution space for both the image downscaling and SR is implicitly constrained. Also, normalizing flow-based generative models are suitable for high-quality and diverse image downscaling and SR with stable learning \cite{xiao2020invertible,liang2021hierarchical}. Next, we first present a preliminary introduction to the normalizing flow, then the model architecture of the SDFlow is elaborated.

\subsubsection{Normalizing flow}
Normalizing flow \cite{kobyzev2020normalizing}, as a type of latent variable generative model, is to explicitly estimate the probability distribution of the input data. Compared to GAN \cite{goodfellow2014generative} and variational autoencoder (VAE) \cite{Kingma2014}, normalizing flow features exact likelihood estimation, efficient sample generation, and stable training. Typically, flow-based models try to learn a bijective mapping between the data space and the latent space. Specifically, let $(\mathbf{x}\in \mathbb{R}^d) \sim p(\mathbf{X})$ be a $d$ dimensional random variable with complex marginal distribution, the main idea of flow-based models is the employment of a learned invertible function $f_\theta(\cdot)$ transforming $\mathbf{x}$ to $(\mathbf{z}\in \mathbb{R}^d) \sim p(\mathbf{Z})$ with simple and tractable distribution (\eg~ Gaussian distribution). Since $f_\theta(\cdot)$ is invertible, it implies a one-to-one mapping between the input and output. Thus, the data can be losslessly restored from the latent variable as $\mathbf{x}=f_\theta^{-1}(\mathbf{z})$. Making the use of change-of-variables rule, the log-likelihood the of data $\mathbf{x}$ can be written as:
\begin{equation}
    \log p(\mathbf{x}) = \log p_\phi(\mathbf{z}=f_\theta(\mathbf{x})) + \log \left| \text{Det} \frac{\partial f_\theta(\mathbf{x})}{\partial \mathbf{x}} \right|
\end{equation}
where $\phi$ is the parameter of the base distribution, \eg~ for gaussian distribution $\phi$ represents the mean and variance. $\log \left| \text{Det} \frac{\partial f_\theta(\mathbf{x})}{\partial \mathbf{x}} \right|$ is the logarithm of the absolute value of the determinant Jacobian of $f_\theta(\mathbf{x})$ with respect to $\mathbf{x}$.

The main challenge in designing a flow-based model for practical use is to ensure that $f_\theta(\cdot)$ is expressive enough, while the log-determinant of the Jacobian of $f_\theta(\cdot)$ must be easy and efficient to compute. Usually, $f_\theta(\cdot)$ is composed of a stack of invertible functions $f_\theta^1\circ f_\theta^2\circ f_\theta^3\circ\cdot\cdot\cdot\circ f_\theta^K$ , as all together they still represent a single invertible function. The intermediate variables are defined as $\mathbf{h}^k=f_\theta^k(\mathbf{h}^{k-1})$, where $\mathbf{h}^0$ and $\mathbf{h}^k$ represent the input random variable $\mathbf{x}$ and the output latent random variable $\mathbf{z}$. Using multiple learnable invertible functions, a normalizing flow attempts to transform the input random variable from the data space into a latent variable obeying a simple distribution. Thus, the log-likelihood of the input variable can be finally formulated as:
\begin{equation}\label{eq:nf}
    \log p(\mathbf{x}) = \log p_\phi(\mathbf{z}) + \sum_{k=1}^K \log \left| \text{Det} \frac{\partial f_\theta^k(\mathbf{\mathbf{h}^{k-1}})}{\partial \mathbf{h}^{k-1}} \right|
\end{equation}
generally, the individual invertible function $f_\theta^k(\cdot)$ is parameterized by neural networks, and the entire flow model parameters can be optimised by minimizing the exact negative log-likelihood (NLL) on the training dataset using standard stochastic gradient descent optimization method.

One caveat in designing the INN for normalizing flow is to ensure that the determinate Jacobian of the high-dimensional transformation can be computationally tractable. Common INN realizations use coupling layers such that the Jacobian is an upper or lower triangular matrix, and the determinant can be easily evaluated as the product of its diagonal elements. Dinh \etal \cite{Laurent2015} build flow model using a stack of non-linear additive coupling layers in which the determinant  is always $1$. Further, the Real-NVP \cite{Laurent2017} enhance the additive coupling to affine coupling that performs both location and scale transformations. Recently, Glow \cite{kingma2018glow} replaces the fixed permutation layer in \cite{Laurent2015,Laurent2017} with the more flexible $1\times 1$ convolutional layer and achieves better image generation quality.

\begin{figure*}
    \centering
    \includegraphics[width=\textwidth]{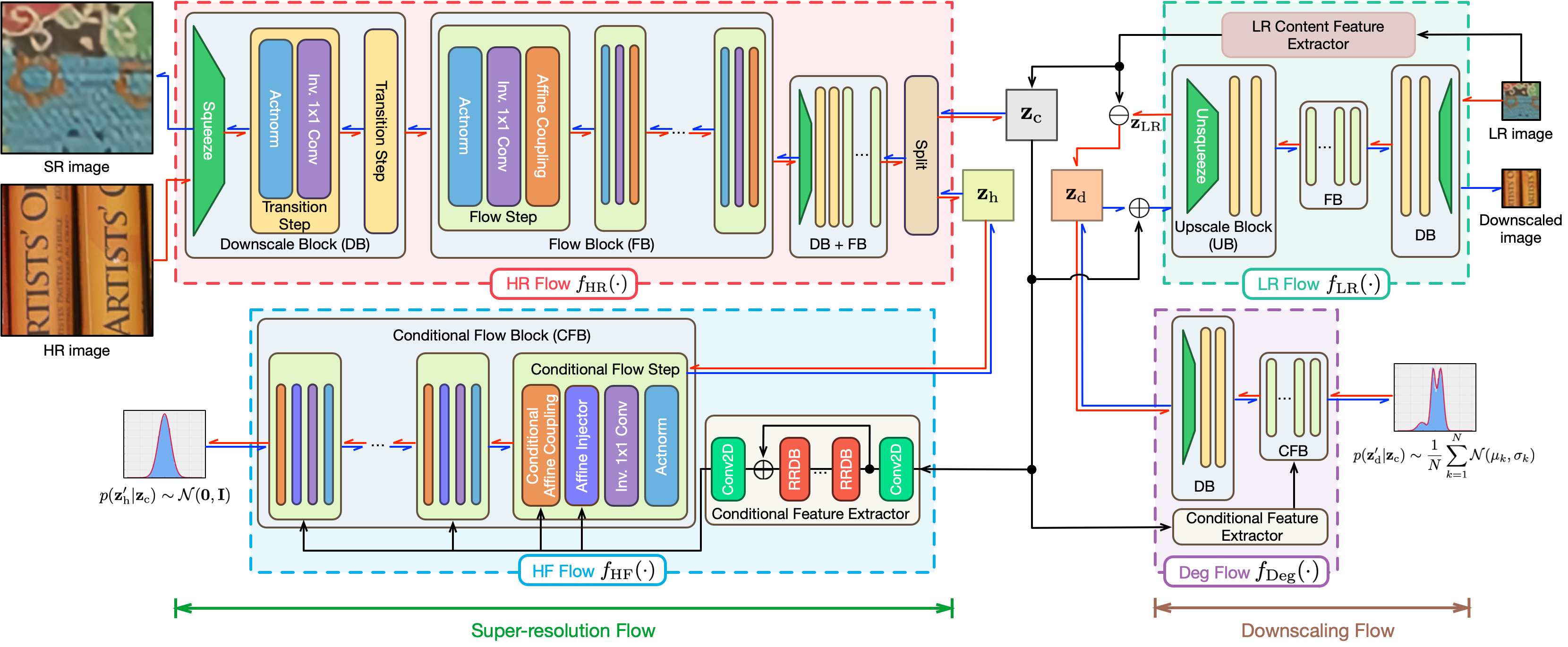}
    \caption{Overview of our proposed SDFlow framework. In the forward pass, HR Flow maps the HR input into the latent space and divides the latent variable into $[\mathbf{z}_\text{c}, \mathbf{z}_\text{h}]$, representing the image content and the HF components. The HF Flow projects $\mathbf{z}_\text{h}$ to standard normal distributed $\mathbf{z}'_\text{h}$ under the conditional feature extracted from $\mathbf{z}_\text{c}$.  LR Flow decouples the content $\mathbf{z}_\text{c}$ and degradation $\mathbf{z}_\text{d}$ of the input LR image in the latent space at the forward pass. Then the Deg Flow further transforms $\mathbf{z}_\text{d}$  into $\mathbf{z}'_\text{d}$ under the conditional features extracted from $\mathbf{z}_\text{c}$ such that $\mathbf{z}'_\text{d}$ follows the standard normal distributions. Conversely, the sampled HF and content variable of the LR image are taken into the reverse of HF Flow and HR Flow to generate the SR image, while the sampled degradation information and content variable of the HR image are used to generate the downscaled image during the backward pass of the Deg Flow and LR Flow.}
    \label{fig:overview}
\end{figure*}
\subsubsection{Model architecture}\label{sec:model_architecture}
The flow-based models are naturally and intuitively for image SR and downscaling tasks \cite{xiao2020invertible,liang2021hierarchical}, since image SR and downscaling are inherently a mutual inverse process.  In this paper, we propose the SDFlow framework to simultaneously model the SR and downscaling for real-world images in a unified architecture which can fully exploit the underlying complex dependencies between image SR and downscaling. \fig{\ref{fig:overview}} illustrates the proposed SDFlow framework, it mainly consists of two invertible parts, \ie, the Super-resolution Flow and the Downscaling Flow. Since these two flows are both invertible, thus, the entire framework is also invertible. The Super-resolution Flow includes two sub-flows, the HR Flow takes the HR image $\mathbf{y}$ as input and decomposes it into the content latent variable $\mathbf{z}_\text{c}$ and the remaining HF latent variable $\mathbf{z}_\text{h}$ through a series of invertible flow blocks. Taking $\mathbf{z}_\text{c}$ as a condition, the HF Flow further maps $\mathbf{z}_\text{h}$ to $\mathbf{z}'_\text{h}$ subjected to a simple normalized distribution, such as the isotropic Gaussian distribution. As to the downscaling flow, it also contains two sub-flows, the LR Flow takes the LR image as input and decouples its content and degradation into $\mathbf{z}_\text{c}$ and $\mathbf{z}_\text{d}$ in the latent space. The Deg Flow transforms $\mathbf{z}_\text{d}$ into $\mathbf{z}'_\text{d}$ under the conditional features extracted from $\mathbf{z}_\text{c}$ such that $\mathbf{z}'_\text{d}$ follows the standard normal distributions. By virtue of the invertible design, image SR can be achieved by taking $\mathbf{z}_\text{c}$ of the LR image and the sampled $\mathbf{z}'_\text{h}$ into the inverse process of the Super-resolution Flow to generate one possible SR image. Similarly, an instance of the downscaled images can be obtained by transforming $\mathbf{z}_\text{c}$ of the HR image and the sampled $\mathbf{z}'_\text{d}$ using the inverse process of the Downscaling Flow. The entire model parameters are optimized with the objective function described in \eq{\ref{eq:final_elbo}}. In unpaired learning settings, the latent variable $\mathbf{z}_\text{c}$ indicates a shared latent space between the HR and LR images. In the SDFlow model, we assume that $\mathbf{z}_\text{c}$ captures the image content and the shared latent space is learned using the adversarial training strategy.

\textbf{Super-resolution Flow}: The detailed architecture of the Super-resolution Flow is presented in \fig{\ref{fig:overview}}. The HR Flow model is basically designed upon the unconditional Glow \cite{kingma2018glow} and RealNVP \cite{Laurent2017} models. It is a $L=\log_2(s)$ level architecture, where $s$ is the SR factor. Each level is composed of a downscaling block (DB) and a flow block (FB). Concretely, the DB starts with a checkboard squeeze operation \cite{Laurent2017,lugmayr2020srflow} transforming its input with size $H\times W\times C$ into $\frac{H}{2}\times \frac{W}{2}\times 4C$, then two transition steps \cite{lugmayr2020srflow} are followed to learn a linear invertible interpolation between neighboring elements. Each transition step consists of an invertible activation normalization (\texttt{ActNorm}) operation and an invertible $1\times 1$ convolution (\texttt{Inv. $1\times 1$ Conv}) layer. In each level, the FB consists $K$ flow steps, transforming the input features into the target latent space. Each flow step contains an \texttt{ActNorm}, an \texttt{Inv. $1\times 1$ Conv} and an affine coupling (\texttt{Affine Coupling}).  Here we briefly describe the above mentioned elementary invertible operations, detailed information can be referred to \cite{kingma2018glow,lugmayr2020srflow}:
\begin{itemize}
    \item \texttt{ActNorm}: activation normalization proposed in \cite{kingma2018glow} is an alternative to the batch normalization to alleviate challenges in model training. \texttt{ActNorm} learns a channel-wise scaling and shift to normalize intermediate features.
    \item \texttt{Inv. $1\times 1$ Conv}: to optimize the use of each channel of features between flow steps, each feature channel should have chances to be modified. Instead of using a pre-defined and fix channel permutation strategy \cite{Laurent2015,Laurent2017}, Glow \cite{kingma2018glow} proposed an invertible $1\times 1$ convolution, which generalizes any permutation of channel ordering.
    \item \texttt{Affine Coupling}: affine coupling layer is an efficient invertible transformation that captures complex dependencies between feature channels. It splits the feature into two parts $[\mathbf{z}_a, \mathbf{z}_b]$ along the channel dimension. $\mathbf{z}_a$ is kept unchanged and used to compute the scale and shift to affine transform $\mathbf{z}_b$. Affine coupling is formulated as:
\begin{equation}
    \begin{aligned}
        \mathbf{z}'_a =& \mathbf{z}_a\\
        \mathbf{z}'_b =& \exp(h_\text{s}(\mathbf{z}_a))\mathbf{z}_b + h_\text{b}(\mathbf{z}_a)
    \end{aligned}
\end{equation}
where $h_\text{s}(\cdot)$ and $h_\text{b}(\cdot)$ are arbitrary neural networks. The output of \texttt{Affine Coupling} layer is the concatenated $\mathbf{z}'_a$ and $\mathbf{z}'_b$.
\end{itemize}

At the end of the HR Flow, the transformed latent variable of the input HR image is split into two parts along the channel dimension,\ie, $\mathbf{z}_\text{c}\in \mathbb{R}^{\frac{H}{2^L}\times \frac{W}{2^L}\times 3}$ and $\mathbf{z}_\text{h}\in \mathbb{R}^{\frac{H}{2^L}\times \frac{W}{2^L}\times (2^L\times C - 3)}$. Using the change-of-variable formula, we arrive at the following equation:
\begin{equation}\label{eq:mle_hr}
    p(\mathbf{z}_\text{c}|\mathbf{y})p(\mathbf{z}_\text{h}|\mathbf{z}_\text{c}, \mathbf{y}) = p(\mathbf{z}_\text{c})p(\mathbf{z}_\text{h})\left|\text{Det}\frac{\partial f_\text{HR}(\mathbf{y})}{\partial \mathbf{y}}\right|
\end{equation}
if we consider the HR Flow as a VAE with shared encoder and decoder modelled as Dirac deltas, then the reconstruction term of $\text{ELBO}_{\mathbf{y}}$ in \eq{\ref{eq:final_elbo}} is automatically satisfied with $\mathbf{y}=f_{\text{HR}}^{-1}([\mathbf{z}_\text{c}, \mathbf{z}_\text{h}])$.  Since the invertibility of the transformation of the HR Flow, the two KL divergence terms of $\text{ELBO}_{\mathbf{y}}$ in \eq{\ref{eq:final_elbo}} are equivalent to fitting a flow-based model to the data distribution \cite{George2021normalizing}, which is further equivalent to fitting the HR Flow model to samples $\{\mathbf{y}_i\}_{i=1}^M$ by maximum likelihood estimation (MLE) of \eq{\ref{eq:mle_hr}} \cite{George2021normalizing}.

In typical flow-based models, the factorted out latent variable $\mathbf{z}_\text{c}$ and $\mathbf{z}_\text{h}$ are directly Gaussianized to model the sampling space. However, in the SDFlow model, we do not have to sample from the content latent space to generate SR images, thus, we assume that the prior $p(\mathbf{z}_\text{c})$ is uniform. To model the sampling space of $\mathbf{z}_\text{h}$, direct Gaussianization may lead to suboptimal results due to the lack of sufficient modeling ability of the limited flow depth. Inspired by the conditional flow \cite{liang2021hierarchical}, it is assumed that the HF component relies on the content component to facilitate the distribution modeling of the HF component of natural images. Therefore, $\mathbf{z}_\text{h}$ is further transformed into $\mathbf{z}'_\text{h}=f_\text{HF}(\mathbf{z}_\text{h}, \mathbf{z}_\text{c})$ conditioned on features extracted from $\mathbf{z}_\text{c}$, and $p(\mathbf{z}_\text{h})$ in \eq{\ref{eq:mle_hr}} can be written as $p(\mathbf{z}_\text{h})=p(\mathbf{z}'_\text{h})\left|\text{Det}\frac{\partial f_\text{HF}(\mathbf{z}_\text{h}, \mathbf{z}_\text{c})}{\partial \mathbf{z}_\text{h}}\right|$, where the base distribution of $\mathbf{z}'_\text{h}$ is a multivariate standard normal distribution. As shown in \fig{\ref{fig:overview}}, the HF Flow $f_\text{HF}(\cdot)$ is constructed by $P$ conditional flow steps and a conditional feature extractor. Different from the flow step, conditional flow step replaces the \texttt{Affine Coupling} with \texttt{Conditional Affine Coupling} \cite{lugmayr2020srflow} and introduces the \texttt{Affine Injector} layer. The \texttt{Conditional Affine Coupling} is a conditional variant of the \texttt{Affine Coupling}. The difference is that the scale and shift factors are computed by neural networks on both the input and conditional variables. The \texttt{ affine injector} \cite{lugmayr2020srflow} takes the conditional features to compute the scale and shift factors to transform its input, building a stronger connection between the input and the conditional variables. The conditional feature extractor is a modified version of the RRDB model \cite{esrgan} by adjusting the number of RRDB blocks and removing the upsampling operation. After modeling the distribution of latent space of $\mathbf{z}_\text{c}$ and $\mathbf{z}'_\text{h}$, the MLE of \eq{\ref{eq:mle_hr}} can be optimized by NLL of the following equation:
\begin{equation}\label{eq:nll_hr}
    \begin{aligned}
        NLL_\mathbf{y} &= -\log p(\mathbf{z}'_\text{h}) - \log \left|\text{Det}\frac{\partial f_\text{HR}(\mathbf{y})}{\partial \mathbf{y}}\right|\\
        &- \log \left|\text{Det}\frac{\partial f_\text{HF}(\mathbf{z}_\text{h}, \mathbf{z}_\text{c})}{\partial \mathbf{z}_\text{h}}\right|
    \end{aligned}
\end{equation}

\textbf{Downscaling Flow}: the design paradigm of the Downscaling Flow is much the same as that of the Super-resolution Flow. One of the most notable differences is how we factor out content representation $\mathbf{z}_\text{c}$ and degradation information $\mathbf{z}_{d}$ using the LR Flow. Due to the limitation of the inherent property of the the INN architecture that the input and output dimension must be equal, the LR Flow cannot directly factor out $\mathbf{z}_\text{c}$ and $\mathbf{z}_\text{d}$ using split operation as that of the HR Flow. Additionally, the invertible requirement restricts the flexibility of known transformations composing the INN \cite{Laurent2015,Laurent2017} and leads to inferior capability in disentangling content and degradation. We thus propose a factor out scheme which replaces the split operation with the element-wise minus (plus) operation. The architecture of the LR Flow is shown in \fig{\ref{fig:overview}}, it is composed of a LR content feature extractor and an INN. The LR content feature extractor is a convolution-based neural network extracting content feature $\mathbf{z}_\text{c}$ from the input LR image. The LR image representation $\mathbf{z}_\text{LR}$ in the latent space can be obtained using the INN for its nature of one-to-one mapping. Therefore, the degradation component can be obtained as $\mathbf{z}_\text{d}=\mathbf{z}_\text{LR} - \mathbf{z}_\text{c}$. LR Flow is still invertible as $\mathbf{z}_\text{LR} = \mathbf{z}_\text{c} + \mathbf{z}_\text{d}$.

\begin{figure}
    \centering
    \includegraphics[width=\columnwidth]{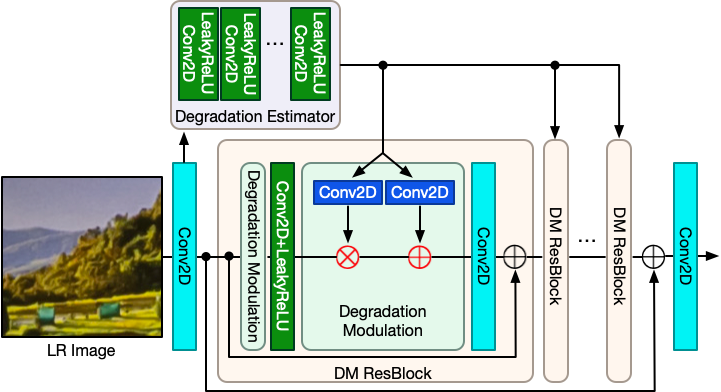}
    \caption{Architecture of the LR Content Feature Extractor. It is designed based on the Resnet where Degradation Modulation is added to learn degradation adaptive transformation.}
    \label{fig:content_extractor}
\end{figure}

\fig{\ref{fig:content_extractor}} illustrates the architecture of the LR content feature extractor. It is basically a residual neural network \cite{resnet} where the ResBlock is replaced with the DM ResBlock. Different from the formal ResBlock, two \texttt{Degradation Modulation} layers are added to do spatial feature affine transformation conditioned on degradation features extracted by the degradation estimator. The degradation estimator contains a series of \texttt{Conv2D} and \texttt{LeakyReLU} layers extracting spatially degradation information from real-world LR images.

The INN in the LR Flow is composed of a DB, a FB and a Upscale Block (UB). Let the input LR image size be $H'\times W'\times 3$, where $H'=\frac{H}{s}$ and $W'=\frac{W}{s}$. The DB first transforms the input LR image into a feature space with size $\frac{H'}{2}\times \frac{W'}{2}\times 12$ to better exploit the correlations between different feature channels, since splitting features with 3 channels into two parts can lead to information deficiency in the \texttt{Affine Coupling} layers in the following FB \cite{zhang2023invertible}. Then the downscaled features are transformed by $K$ flow steps in the FB. Finally, an UB unsqueezes the transformed features into the latent space with the same size as the input LR image.

The forward pass of the LR Flow outputs two latent variables, \ie, $\mathbf{z}_\text{c}\in \mathbb{R}^{H'\times W'\times 3}$ and $\mathbf{z}_\text{d}\in \mathbb{R}^{H'\times W'\times 3}$ of the input LR image. Using the change-of-variables formula, we can get the following equation:
\begin{equation}\label{eq:mle_lr}
    \begin{aligned}
        p(\mathbf{z}_\text{c}|\mathbf{x})p(\mathbf{z}_\mathbf{d}|\mathbf{z}_\text{c}, \mathbf{x}) =& p(\mathbf{z}_\text{c})p(\mathbf{z}_\text{d})\left|\text{Det}\frac{\partial f_\text{LR}(\mathbf{x})}{\partial\mathbf{x}} \right|\left|\text{Det}\frac{\partial \mathbf{z}_\text{d}}{\partial\mathbf{z}_\text{LR}} \right|\\
        =& p(\mathbf{z}_\text{c})p(\mathbf{z}_\text{d})\left|\text{Det}\frac{\partial f_\text{LR}(\mathbf{x})}{\partial\mathbf{x}} \right|
    \end{aligned}
\end{equation}
where $\mathbf{z}_\text{d}=\mathbf{z}_\text{LR}-\mathbf{z}_\text{c}$. Following the derivation of the optimization objective for the HR Flow, the reconstruction term in $\text{ELBO}_\mathbf{x}$ of \eq{\ref{eq:final_elbo}} is automatically satisfied with $\mathbf{x}=f^{-1}_\text{LR}(\mathbf{z}_\text{c}, \mathbf{z}_\text{d})$ and the optimization of the two KL terms of $\text{ELBO}_\mathbf{x}$ is equivalent to the MLE of \eq{\ref{eq:mle_lr}}.

Akin to the Super-resolution Flow, we impose the uniform prior on the content latent variable, since we do not have to sample from the content latent space. To model the sample space of $\mathbf{z}_\text{d}$, we further transform $\mathbf{z}_\text{d}$ into $\mathbf{z}'_\text{d}$ using an additional conditional flow network $f_\text{Deg}(\cdot)$ conditioned on features extracted from $\mathbf{z}_\text{c}$. The structure of the Deg Flow resembles that of the HF Flow. Apart from a conditional feature extractor and $P$ conditional flow steps, a DB is employed to make full use of correlations between different feature channels. After the transformation of the Deg Flow, $p(\mathbf{z}_\text{d})$ of \eq{\ref{eq:mle_lr}} can be written as $p(\mathbf{z}_\text{d})=p(\mathbf{z}'_\text{d})\left|\text{Det}\frac{\partial f_\text{Deg}(\mathbf{z}_\text{d},\mathbf{z}_\text{c})}{\partial \mathbf{z}_\text{d}} \right|$. Considering that real-world image degradations are stochastic and complex, we assume that the base distribution of $\mathbf{z}'_\text{d}$ is a mixture of Gaussians with each representing a potential type of degradation. Thus, $p(\mathbf{z}'_\text{d})$ can be written as $\frac{1}{N}\mathcal{N}(\mathbf{z}'_\text{d}|\mu_i,\Sigma_i)$, where $N$ is the number of Gaussians, $\mu_i$ and $\Sigma_i$ are learnable mean and covariance. After modeling the distribution of the latent space of $\mathbf{z}_\text{c}$ and $\mathbf{z}'_\text{d}$, the MLE of \eq{\ref{eq:mle_lr}} can be optimized by NLL of the following equation:
\begin{equation}\label{eq:nll_lr}
    \begin{aligned}
        NLL_\mathbf{x}&=-\log p(\mathbf{z}'_\text{d})-\log \left|\text{Det}\frac{\partial f_\text{LR}(\mathbf{x})}{\partial \mathbf{x}} \right|\\
        &-\log \left|\text{Det}\frac{\partial f_\text{Deg}(\mathbf{z}_\text{d}, \mathbf{z}_\text{c})}{\partial \mathbf{z}_\text{d}} \right|
    \end{aligned}
\end{equation}

\textbf{Modeling the shared latent space of the content variable}: one key of our proposed method is to meet the assumption that the HR Flow and LR Flow can extract the HR and LR image contents and map them into the same latent space. However, only optimizing \eq{\ref{eq:nll_hr}} and \eq{\ref{eq:nll_lr}} cannot achieve many-to-many mapping between image downscaling and SR, first because it is a nontrivial task for the HR Flow and LR Flow to disentangle content and respective HF and degradation components without any supervision. Additionally, the latent space of HR image content and LR image content are also not well aligned. To address the first issue, we need to constrain that $\mathbf{z}_\text{c}$ of both HR and LR image can be mapped into a valid image capturing the major information of HR and LR inputs. This can be achieved using the inverse process of the LR Flow to map the content latent variable to the LR image space. Inspired by \cite{wolf2021deflow,son2021toward}, we assume that the image content and low-level structure are mainly embodied in the low-frequency part. The following content loss is applied to constrain the content information captured in $\mathbf{z}_\text{c}$ to be consistent with that of the LR and HR image.
\begin{equation}\label{eq:lr_content}
\begin{aligned}
    \mathcal{L}_\text{content} &= ||LPF(f^{-1}_\text{LR}(\mathbf{z}^\text{HR}_\text{c}, 0))-LPF(BD_s(\mathbf{y}))||_1\\
    &+||LPF(f^{-1}_\text{LR}(\mathbf{z}^\text{LR}_\text{c}, 0))-LPF(\mathbf{x})||_1\\
    &+\alpha||\phi(f^{-1}_\text{LR}(\mathbf{z}^\text{HR}_\text{c}, 0))-\phi(BD_s(\mathbf{y}))||_1\\
    &+\alpha||\phi(f^{-1}_\text{LR}(\mathbf{z}^\text{LR}_\text{c}, 0))-\phi(\mathbf{x})||_1
\end{aligned}
\end{equation}
where $LPF(\cdot)$ is a low-pass filter, $BD_s(\cdot)$ is the Bicubic downscaling operation with scale $s$, $\phi(\cdot)$ denotes the pre-trained VGG-19 \cite{Simonyan15} feature extractor. We minimize the feature difference from \texttt{Conv5\_4} of the pre-trained VGG-19 to encourage $\mathbf{z}_\text{c}$ to maintain the texture content \cite{srgan}. The pixel and feature loss are weighted by the factor $\alpha$.

With \eq{\ref{eq:lr_content}}, the training of the Super-resolution Flow and Downscaling Flow are correlated through a shared image decoder, \ie, $f^{-1}_\text{LR}(\cdot)$, which ensures that $\mathbf{z}_\text{c}$ of the HR and LR image are mapped to the same latent space while still encode the image content information. However, domain gap still exists between content latent variables of the HR and LR images. To reduce the marginal distribution difference between two domains, we employ the adversarial domain alignment strategy. Specifically, we train a discriminator differentiating $\mathbf{z}^\text{HR}_\text{c}$ and $\mathbf{z}^\text{LR}_\text{c}$. The following loss function is applied to train the domain discriminator:
\begin{equation}\label{eq:domain}
    \begin{aligned}
        \mathcal{L}_\text{domain}&=||D(\mathbf{z}^\text{HR}_\text{c})||_2 + ||1-D(\mathbf{z}^\text{LR}_\text{c})||_2\\
        &+ \beta_1(||\mathbf{z}^\text{HR}_\text{c}||_2 + ||\mathbf{z}^\text{LR}_\text{c}||_2) + \beta_2||\mathbf{z}_\text{LR} - \text{sg}(\mathbf{z}^\text{LR}_\text{c})||_2
    \end{aligned}
\end{equation}
where $D(\cdot)$ represents the patch discriminator, $\text{sg}(\cdot)$ means stop gradient back-propagation. We use the LSGAN \cite{lsgan} to optimize the parameters of the discriminator, meanwhile parameters of the HR Flow and LR Flow are optimized to fool the discriminator with contradictory LSGAN loss. To ease and stabilize the training of the discriminator, we apply $L_2$ regularization on the content latent variable, which avoids arbitrary high variance latent space and enforces a closer domain gap in a denser latent space.

\subsubsection{Training objectives}
To train the proposed SDFlow model, it is statistically sufficient asymptotically to optimize MLE based loss function as \eq{\ref{eq:nll_hr}} and \eq{\ref{eq:nll_lr}} together with shared content latent space loss as \eq{\ref{eq:lr_content}} and \eq{\ref{eq:domain}} from the statistical perspective. While, in training the generative models, adversarial learning outperforms MLE on sample quality metrics \cite{grover2018flow}. Additionally, the optimality of MLE holds only when there is no model miss-specification for the generative model \cite{ardizzone2018analyzing,grover2018flow}. Fortunately, INN offers the opportunity to simultaneously optimize for losses on both directions, which allows for more effective training. We thus perform distribution matching on the generation side of Super-resolution and Downscaling Flow with the commonly used GAN loss and perceptual loss. The overall training objective function is formulated as follows:
\begin{equation}\label{eq:final_loss}
    \begin{gathered}
        \mathcal{L} = \underbrace{NLL_\mathbf{x} + NLL_\mathbf{y} + \mathcal{L}_\text{content} + \mathcal{L}_\text{domain}}_\text{Forward loss}\\
        \underbrace{\begin{aligned}&+(\lambda_1\mathcal{L}^\text{DS}_\text{pix.} + \lambda_2\mathcal{L}^\text{DS}_\text{per.} + \lambda_3\mathcal{L}^\text{DS}_\text{GAN})\\
        &+ (\lambda_4\mathcal{L}^\text{SR}_\text{pix.} + \lambda_5\mathcal{L}^\text{SR}_\text{per.} + \lambda_6\mathcal{L}^\text{SR}_\text{GAN})\end{aligned}}_\text{Backward loss}
    \end{gathered}
\end{equation}
where $\mathcal{L}^\text{DS}_\text{pix.}$, $\mathcal{L}^\text{DS}_\text{per.}$ and $\mathcal{L}^\text{DS}_\text{GAN}$ are per-pixel loss, perceptual loss and GAN loss applied on the backward pass of the Downscaling Flow, $\mathcal{L}^\text{SR}_\text{pix.}$, $\mathcal{L}^\text{SR}_\text{per.}$ and $\mathcal{L}^\text{SR}_\text{GAN}$ are per-pixel loss, perceptual loss and GAN loss applied on the backward pass of the Super-resolution Flow. Perceptual loss is implemented as the $L_1$ loss applied on features extracted from \texttt{Conv5\_4} layer of the pre-trained VGG-19. GAN loss is realized by adding discriminators on backward side of the Super-resolution and Downscaling Flow, and LSGAN is utilized to optimized the two flows and discriminators. The loss terms in the backward loss are weighted by $\lambda_1$ to $\lambda_6$, respectively. To optimize \eq{\ref{eq:final_loss}}, we perform forward and backward passes of the SDFlow model in an alternating manner, and gradients are accumulated before the parameter update.
	
\section{Experiments}
In this section, we evaluate the performance of the SDFlow in generating diverse SR and downscaled images from real-world LR and HR inputs, respectively. We first introduce experimental setup including real-world image SR datasets, evaluation metrics, and implementation details. Then, experimental results of SR and downscaling are evaluated quantitatively and qualitatively. Finally, ablation studies are conducted to validate the effectiveness of the proposed architecture.

\subsection{Experimental setup}
\subsubsection{Datasets and evaluation metrics}
The SDFlow model is trained on unpaired real-world LR and HR image datasets and evaluated on the corresponding test real-world HR and LR images. We conduct experiments on the widely used RealSR \cite{cai2019toward} and DRealSR \cite{wei2020component} datasets that are originally collected for the real-world image SR task. RealSR dataset has 595 (500 for $\times 4$ SR) pairs of HR and LR image captured by adjusting focal length through Canon and Nikon DSLR cameras. Progressive image registration framework is proposed in order to achieve pixel wise registration of images captured at 28mm, 35mm, 50mm and 105mm. Images captured at 105mm are used as HR ground-truth while images captured at the other three focal lengths are considered as LR counterparts with different scale factors. DRealSR dataset is captured similarly as the RealSR dataset but has a larger scale of image pairs. 5 different DSLR cameras (\ie, Sony, Canon, Olympus, Nikon, and Panasonic) are used to capture indoor and outdoor images at four different resolutions, obtaining 884 (for $\times 2$ SR), 783 (for $\times 3$ SR), 840 (for $\times 4$ SR) LR and HR image pairs. SIFT algorithm is used to align these image pairs.

To evaluate the performance of image SR and downscaling models, Peak Signal-to-Noise Ratio (PSNR) and Structural Similarity Index Measure (SSIM) \cite{ssim} are two widely used image distortion measurement metrics. Following the setting in \cite{xiao2020invertible,liang2021hierarchical}, we evaluate the performance of the proposed SDFlow and other compared models using PSNR and SSIM on the Y channel of the YCbCr color space. However, PSNR and SSIM are only based on local image differences, failing to consider image quality in terms of human perception. Following \cite{chen2022real}, we also employ two full-reference image quality assessment metrics, Learned Perceptual Image Patch Similarity (LPIPS) \cite{lpips}, Information Fidelity Criterion (IFC) \cite{ifc}, and three no-reference image quality assessment metrics, Natural Image Quality Evaluator (NIQE) \cite{niqe}, Perception-based Image Quality Evaluator (PIQE) \cite{piqe} and No-Reference Quality Metric (NRQM) \cite{nrqm} for better visual quality comparison. 10 samples are used to measure the diversity\cite{diversity} of the model outputs.

\subsubsection{Implementation details}
In the implementation of the proposed SDFlow, we set the number of flow steps $K$ and number of conditional flow steps $P$ to 16 and 8, respectively. Regarding the conditional feature extractor, since modeling the solution space of the HF component is more complex than that of the degradation component in terms of the dimensionality of unknown variables, we thus set the number of RRDB to 8 and 4 to extract conditional features for the HF Flow and Deg Flow. In the LR content feature extractor, we use 8 \texttt{Conv2D-LeakyReLU} operations to extract degradation features and use 16 DM ResBlocks to obtain the LR image content feature. The number of Gaussian components are empirically set to 16.

The proposed SDFlow is trained in the training set and tested in the test set of the RealSR and DRealSR dataset, respectively. The weight $\alpha$ in $\mathcal{L}_\text{content}$ is set to 0.05, $\beta_1$ and $\beta_2$ in $\mathcal{L}_\text{domain}$ is set to 0.05 and 0.5, $\lambda_1$ to $\lambda_6$ in the overall loss \eq{\ref{eq:final_loss}} are set to 0.5, 0.5, 0.1, 0.5, 0.5, 0.1, respectively. We use Adam optimizer to update parameters of the SDFlow. The learning rate for the SDFlow model and all of the discriminators are initialized to $1\times 10^{-4}$ and $1\times 10^{-5}$, respectively, and halved at $50\%$, $75\%$, $90\%$, $95\%$ of the total training iterations. We set the mini-batch size to 32 and randomly crop the input HR and LR image into $192\times 192$ and $48\times 48$ patches. Training samples are augmented by applying random horizontal and vertical flip. When composing unpaired mini-batch samples, we explicitly avoid cropping HR and LR patches from paired HR and LR data. Since flow-based models are defined over continuous variables, we dequantize the discrete pixel values by adding uniform noise in range $[0, \frac{1}{255}]$  \cite{flow++}. Due to the fact that adversarial training is used, we first stabilize the training by pretraining the SDFlow with $NLL_\mathbf{x}$, $NLL_\mathbf{y}$ and $\mathcal{L}_\text{content}$ for 50k iterations. The entire model is then trained with the forward loss in \eq{\ref{eq:final_loss}} for 200k iterations. Finally, we fine-tune the SDFlow with forward and backward loss as \eq{\ref{eq:final_loss}} for another 50k iterations. Following the setting in \cite{lugmayr2020srflow,liang2021hierarchical}, we set the sampling temperature $\tau$ to 0 for per-pixel loss and 0.8 for perceptual and GAN loss when optimizing the model with the backward.

\subsection{Evaluation of the SR results}
This section reports the quantitative and qualitative performance of the proposed SDFlow and other state-of-the-art unsupervised RSISR methods. As concluded in \cite{chen2022real}, unsupervised RSISR methods can be categorized into degradation modelling based models, domain translation based models, and self-learning based models. In unpaired SISR settings, degradation modelling based models focus on the recovery of degradation kernels that are then used to synthesize LR images from HR inputs for the following supervised SISR models training. Domain translation based models usually employ a learned CNN to translate HR images (or BD downscaled images) to real-world LR image domain, which generates paired LR/HR images to train generic SISR models. Self-learning based models exploit cross-scale internal recurrence of LR image to learn image-specific degradation model, then test LR image and its degraded counterpart are used to train an image-specific SR model. We compared 10 unpaired RSISR models from the three categories with source code available, \ie, KernelGAN \cite{bell2019blind}, Impressionism \cite{ji2020real}, ADL \cite{son2021toward}, DASR \cite{wei2021unsupervised}, USR-DA \cite{usr-da}, Pseudo-supervision \cite{maeda2020unpaired}, SRResCGAN \cite{umer2020deep}, ZSSR \cite{shocher2018zero}, DBPI \cite{kim2020dual}, DualSR \cite{emad2021dualsr}. All models are re-trained on the RealSR and DRealSR dataset using the training script and default settings provided by the source code. Additionally, Bicubic upscaled LR images are also evaluated for reference.

\begin{table*}[h]
\centering
\caption{Quantitative comparison for $4\times$ SR on the RealSR test dataset. $\uparrow$ indicates the higher score the better performance, while $\downarrow$ means the lower score the better performance.}\label{tab:sr_realsr}
\resizebox{\textwidth}{!}{\begin{tabular}{c|c|ccccc|ccc} 
\hline
\multirow{2}{*}{}                                                               & \multirow{2}{*}{Methods}               & \multicolumn{5}{c|}{Full-reference}                                                            & \multicolumn{3}{c}{No-reference}                       \\ 
\cline{3-10}
                                                                                &                                        & PSNR$\uparrow$   & SSIM$\uparrow$  & LPIPS$\downarrow$ & IFC$\uparrow$   & Diversity$\uparrow$ & NIQE$\downarrow$ & PIQE$\downarrow$ & NRQM$\uparrow$   \\ 
\hline\hline
Interpolation                                                                   & Bicubic                                & 27.2334          & 0.7637          & 0.4761            & 0.9632          & -                   & 8.9289           & 92.3236          & 2.7284           \\ 
\hline
\multirow{3}{*}{\begin{tabular}[c]{@{}c@{}}Degradation\\modelling\end{tabular}} & KernelGAN                              & 25.1433          & 0.7387          & 0.3366            & 0.9542          & -                   & 6.9917           & 80.5237          & 3.7027           \\
                                                                                & Impressionism                          & 25.2184          & 0.7025          & 0.3608            & 0.9183          & -                   & \textbf{4.0966}  & \textbf{35.9179} & \textbf{6.8024}  \\
                                                                                & ADL                                    & 28.3478          & 0.8102          & 0.3138            & 1.1916          & -                   & 8.0719           & 85.8044          & 3.3137           \\ 
\hline
\multirow{4}{*}{\begin{tabular}[c]{@{}c@{}}Domain\\translation\end{tabular}}    & DASR                                   & 27.4852          & 0.7872          & 0.2646            & 1.0658          & -                   & 6.6091           & 66.2664          & 4.2414           \\
                                                                                & USR-DA                                 & 27.3762          & 0.7569          & 0.2683            & 0.9667          & -                   & 5.1228           & 49.9268          & 5.1042           \\
                                                                                & Pseudo-supervision                     & 27.3066          & 0.7750          & 0.2626            & 1.0700          & -                   & 6.8147           & 78.0688          & 4.3592           \\
                                                                                & SRResCGAN                              & 26.7526          & 0.7487          & 0.2829            & 0.9759          & -                   & 5.6330           & 58.9674          & 4.4693           \\ 
\hline
\multirow{3}{*}{\begin{tabular}[c]{@{}c@{}}Self\\learning\end{tabular}}         & ZSSR                                   & 27.5745          & 0.7815          & 0.3845            & 1.0382          & -                   & 7.9022           & 85.6690          & 3.0763           \\
                                                                                & DBPI                                   & 23.8786          & 0.6970          & 0.3170            & 0.9409          & -                   & 6.6047           & 71.1269          & 4.5050           \\
                                                                                & DualSR                                 & 22.5764          & 0.7037          & 0.3582            & 0.8876          & -                   & 5.5278           & 69.9079          & 5.4125           \\ 
\hline
\multirow{4}{*}{SDFlow}                                                         & Forward loss ($\tau = 0$)              & \textbf{28.5813} & \textbf{0.8118} & 0.2934            & \textbf{1.2084} & -                   & 7.9582           & 86.1925          & 3.4795           \\
                                                                                & Forward loss ($\tau = 0.8$)            & 27.8403          & 0.7819          & 0.2321            & 1.0448          & 11.9467             & 5.7145           & 55.3054          & 4.6572           \\
                                                                                & Forward + Backward loss ($\tau = 0$)   & 28.0020          & 0.7937          & 0.2500            & 1.1180          & -                   & 6.9344           & 75.7292          & 4.0850           \\
                                                                                & Forward + Backward loss ($\tau = 0.8$) & 27.1591          & 0.7576          & \textbf{0.2295}   & 1.0090          & \textbf{12.0287}    & 4.8106           & 44.6445          & 5.4873           \\
\hline
\end{tabular}}
\end{table*}

\begin{table*}[h]
\centering
\caption{Quantitative comparison for $4\times$ SR on the DRealSR test dataset. $\uparrow$ indicates the higher score the better performance, while $\downarrow$ means the lower score the better performance.}\label{tab:sr_drealsr}
\resizebox{\textwidth}{!}{\begin{tabular}{c|c|ccccc|ccc} 
\hline
\multirow{2}{*}{}                                                               & \multirow{2}{*}{Methods}               & \multicolumn{5}{c|}{Full-reference}                                                            & \multicolumn{3}{c}{No-reference}                       \\ 
\cline{3-10}
                                                                                &                                        & PSNR$\uparrow$   & SSIM$\uparrow$  & LPIPS$\downarrow$ & IFC$\uparrow$   & Diversity$\uparrow$ & NIQE$\downarrow$ & PIQE$\downarrow$ & NRQM$\uparrow$   \\ 
\hline\hline
Interpolation                                                                   & Bicubic                                & 30.5557          & 0.8595          & 0.4387            & 0.8532          & -                   & 10.0721          & 94.4191          & 2.7187           \\ 
\hline
\multirow{3}{*}{\begin{tabular}[c]{@{}c@{}}Degradation\\modelling\end{tabular}} & KernelGAN                              & 28.7268          & 0.8347          & 0.3995            & 0.8207          & -                   & 8.7270           & 88.7390          & 3.0359           \\
                                                                                & Impressionism                          & 28.1383          & 0.7857          & 0.3810            & 0.6900          & -                   & \textbf{4.9480}  & \textbf{35.5378} & \textbf{5.9220}  \\
                                                                                & ADL                                    & 30.8912          & 0.8722          & 0.3574            & 0.9113          & -                   & 9.5483           & 88.7854          & 2.7541           \\ 
\hline
\multirow{4}{*}{\begin{tabular}[c]{@{}c@{}}Domain\\translation\end{tabular}}    & DASR                                   & 29.6621          & 0.8362          & 0.3126            & 0.7428          & -                   & 6.9862           & 64.7885          & 3.7215           \\
                                                                                & USR-DA                                 & 30.0489          & 0.8371          & 0.2966            & 0.7355          & -                   & 6.1054           & 47.6347          & 3.9122           \\
                                                                                & Pseudo-supervision                     & 29.9271          & 0.8461          & 0.2902            & 0.8483          & -                   & 7.7190           & 81.6083          & 3.6742           \\
                                                                                & SRResCGAN                              & 28.8888          & 0.8177          & 0.3178            & 0.7730          & -                   & 6.5440           & 69.4008          & 3.7875           \\ 
\hline
\multirow{3}{*}{\begin{tabular}[c]{@{}c@{}}Self\\learning\end{tabular}}         & ZSSR                                   & 30.6414          & 0.8627          & 0.3830            & 0.8681          & -                   & 9.2237           & 86.1406          & 2.7894           \\
                                                                                & DBPI                                   & 24.6870          & 0.6795          & 0.4296            & 0.6389          & -                   & 7.2807           & 72.0693          & 4.1016           \\
                                                                                & DualSR                                 & 26.2933          & 0.7855          & 0.3522            & 0.7166          & -                   & 7.0007           & 78.0014          & 4.4276           \\ 
\hline
\multirow{4}{*}{SDFlow}                                                         & Forward loss ($\tau = 0$)              & \textbf{31.0747} & \textbf{0.8726} & 0.3411            & \textbf{0.9266} & -                   & 9.2169           & 87.4987          & 2.9638           \\
                                                                                & Forward loss ($\tau = 0.8$)            & 30.5901          & 0.8592          & 0.2852            & 0.8368          & 10.0632             & 7.6900           & 71.2724          & 3.4047           \\
                                                                                & Forward + Backward loss ($\tau = 0$)   & 30.6352          & 0.8609          & 0.3108            & 0.8912          & -                   & 7.8275           & 82.7311          & 3.3114           \\
                                                                                & Forward + Backward loss ($\tau = 0.8$) & 29.4299          & 0.8073          & \textbf{0.2807}   & 0.7419          & \textbf{12.5713}    & 5.5167           & 39.2800          & 4.9034           \\
\hline
\end{tabular}}
\end{table*}

\begin{figure*}
\centering
    \subfloat{\includegraphics[width=0.9\textwidth]{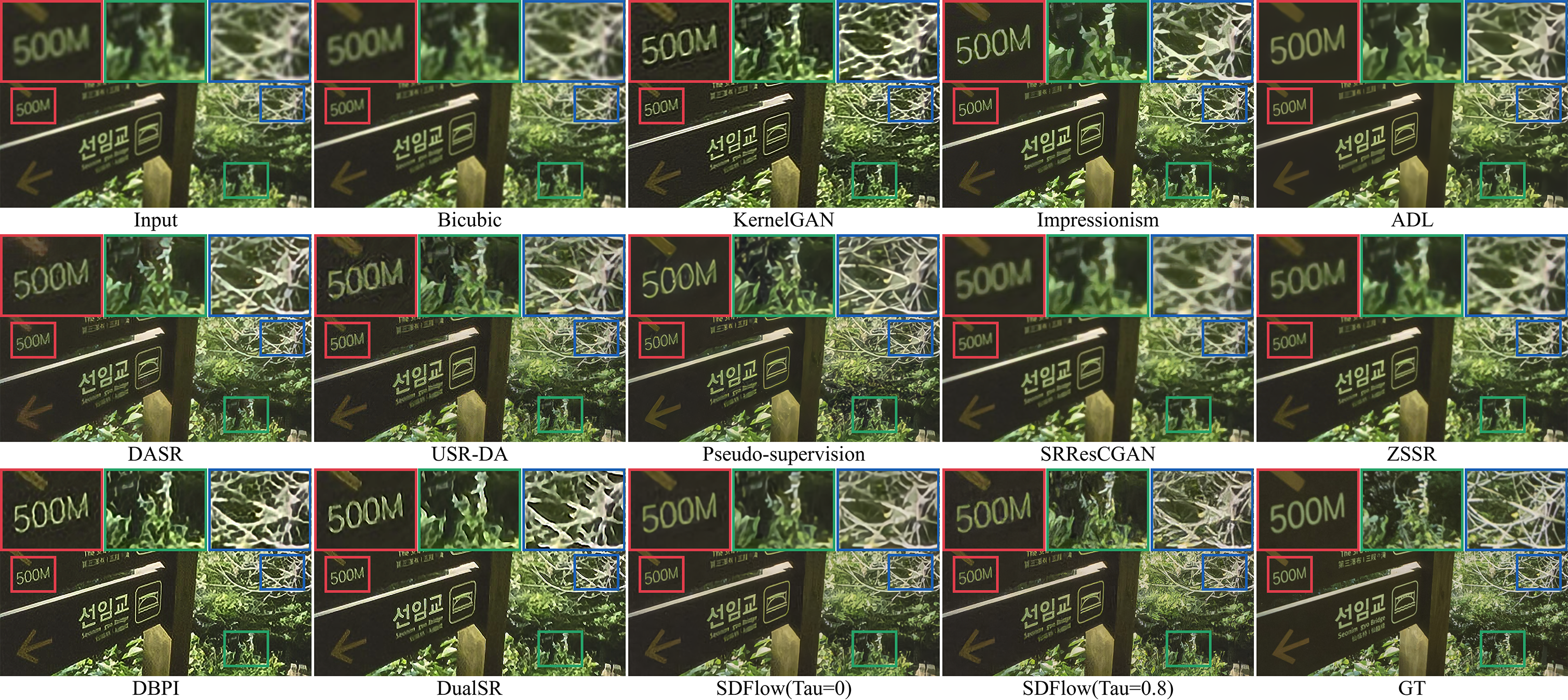}}\\
    \vspace{-10pt}
    \subfloat{\includegraphics[width=0.9\textwidth]{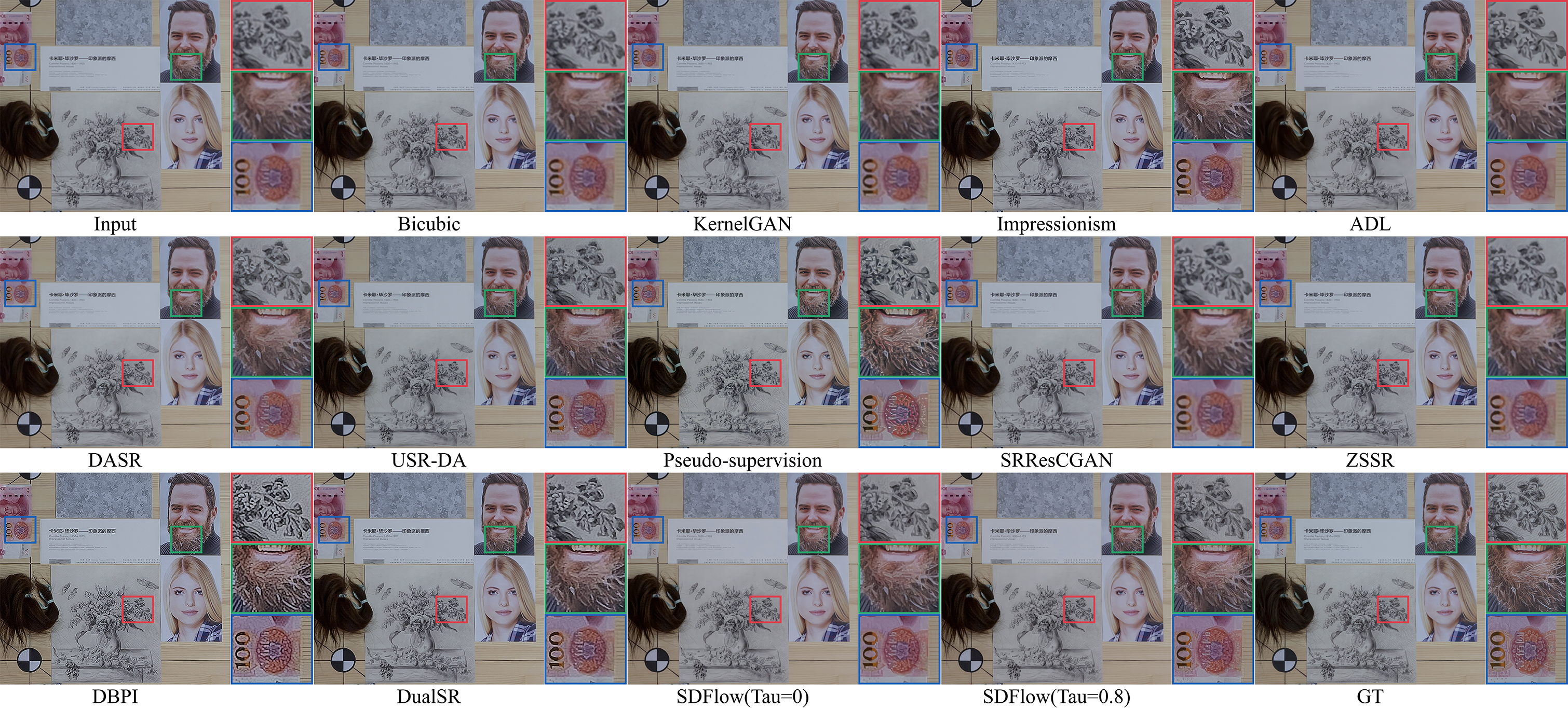}}\\
    \vspace{-10pt}
    \subfloat{\includegraphics[width=0.9\textwidth]{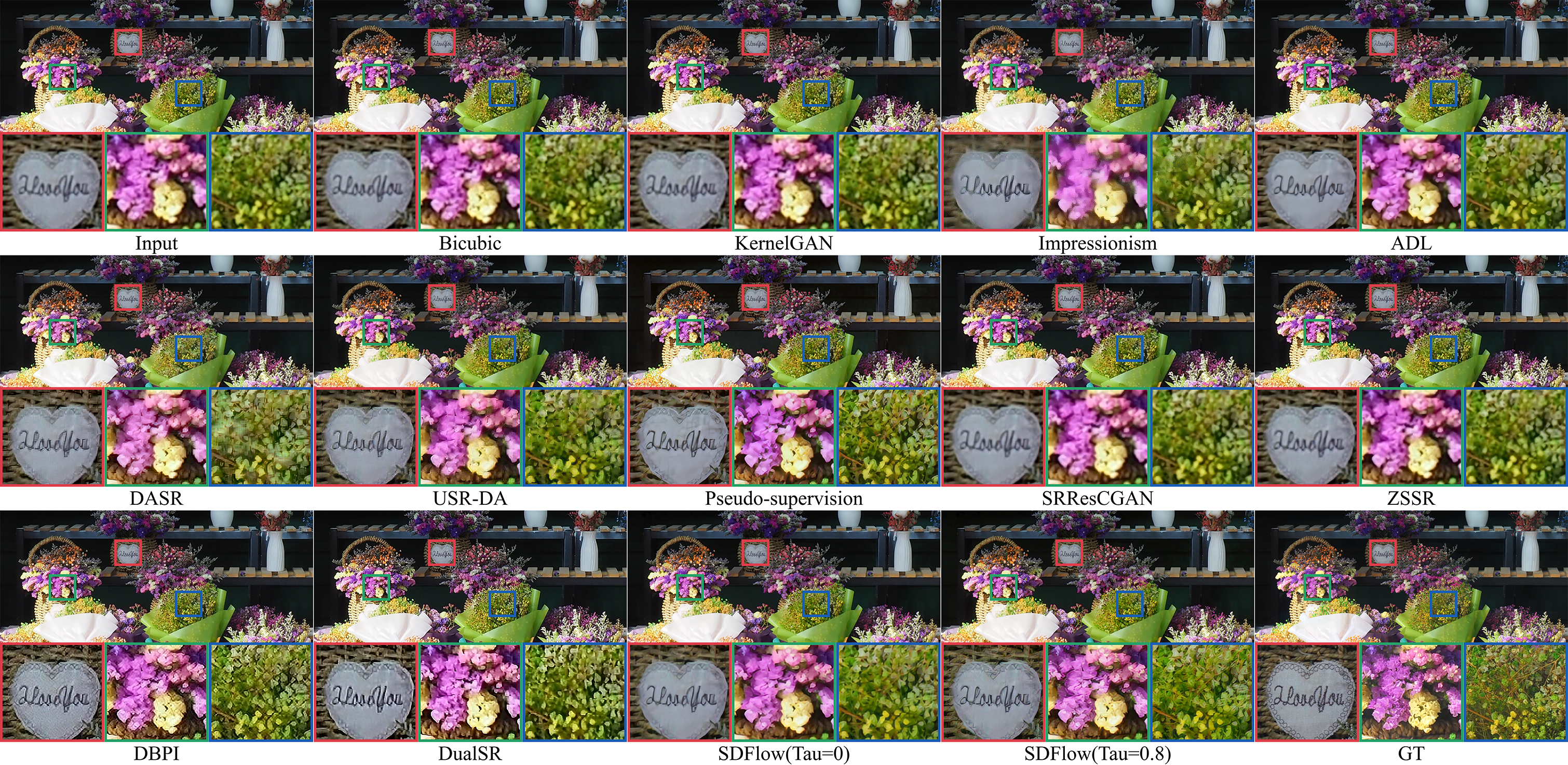}}
    \caption{Qualitative $4\times$ SR images on the RealSR and DRealSR test dataset. Three parts of each SR image are magnified for better visual comparison. The proposed SDFlow can produce better visual pleasing textures.}\label{fig:qualitative_SR}
\end{figure*}

Table \ref{tab:sr_realsr} and \ref{tab:sr_drealsr} present the $4\times$ SR performance of the proposed SDFlow and other compared unsupervised RSISR models on the RealSR and DRealSR dataset, respectively. As can be observed from Table \ref{tab:sr_realsr} and \ref{tab:sr_drealsr}, the proposed SDFlow achieves the best performance on the PSNR, SSIM and IFC metrics with forward loss and $\tau=0$ settings. This is because in this settings, optimization of the SDFlow can be treated as using the pixel-wise L1 loss \cite{Andreas2022}, and deterministic results are generated when sampling temperature is set to 0, thus higher PSNR and SSIM can be achieved. However, higher PSNR and SSIM do not mean that they correlate with better human visual perception, as shown in \fig{\ref{fig:qualitative_SR}}, the generated SR images tend to be blurry. When $\tau=0.8$, the perceptual metrics of the SDFlow are improved significantly. When the SDFlow is trained with both forward and backward loss, the perceptual metrics LPIPS, NIQE, PIQE and NRQM are improved compared with only forward loss in the setting of $\tau=0$, since the incorporation of perceptual and GAN losses in the backward loss term as shown in \eq{\ref{eq:final_loss}}. When $\tau=0.8$, SDFlow obtains the best LPIPS score, and no-reference metrics NIQE, PIQE and NRQM are improved by a large margin. As can be observed, the proposed SDFlow trained with forward and backward loss obtains the second-best performance on the no-reference metrics.

For the compared methods, ZSSR \cite{shocher2018zero}, DBPI \cite{kim2020dual}, DualSR \cite{emad2021dualsr} and KernelGAN \cite{bell2019blind} are all image-specific SISR methods based on the internal learning strategy. However, these methods do not perform well in terms of the full-reference metrics. The primary reason may be that numerous constraints in these methods, \eg, Gaussian kernels and simple kernel shape assumption do not hold for spatial varying stochastic and complex real-world degradations \cite{son2021toward}. Impressionism \cite{ji2020real} adopts KernelGAN to estimate degradation kernels to synthesize HR/LR image pairs for supervised SISR model training, thus inaccurate degradation estimation results in inferior performance on full-reference metrics. However, Impressionism method achieves the best performance on the no-reference metrics. The ADL model \cite{son2021toward} is a two-stage SISR method in which a novel low-pass filter loss and adaptive data loss are proposed to model more accurate degradation kernels to synthesize HR/LR pairs for the second supervised training stage. ADL uses RRDB as the SR backbone and obtains the second best performance on the full-reference metrics. As a comparison, performance of the proposed SDFlow trained with forward loss and $\tau=0$ is superior to that of the ADL on both full-reference and no-reference metrics. Domain translation based models usually train a image downsampler and SR model in a two-stage manner or jointly under the adversarial training framework. These models usually do not obtain high PSNR, SSIM and IFC scores, since PSNR, SSIM and IFC are anti-correlated with the subjective scores, thus are inappropriate for evaluating GAN-based algorithms \cite{jinjin2020pipal}. As pointed by Gu \etal \cite{jinjin2020pipal}, LPIPS correlates mostly with the subjective scores. In Table \ref{tab:sr_realsr} and \ref{tab:sr_drealsr}, most of domain translation based methods get better LPIPS score than that of the other methods been compared. By comparison, the proposed SDFlow model with $\tau=0.8$ achieves better LPIPS, NIQE, PIQE and NRQM scores than that of the compared domain translation based SISR methods.

In addition to the quantitative results, we present qualitative $4\times$ SR images shown in \fig{\ref{fig:qualitative_SR}}, in which local regions with structural content or full of textures are magnified. SR images proposed by the SDFlow model contain clean and perceptual pleasing textures, which are more similar to real HR ground-truth images. As to the other methods been compared, domain translation based DASR, USR-DA and Pseudo-supervision can also produce moderate perceptual friendly results. Impressionism and KernelGAN generate SR images with notable artifacts due to inaccurate degradation estimation. Self-learning based DBPI and DualSR employ adversarial training, which tends to produce SR images with exaggerated textures.

\subsection{Evaluation of the downscaling results}
In this section, we compare LR images downscaled from real-world HR images using the proposed SDFlow with other unsupervised image downscaling (degradation modeling) methods both quantitatively and qualitatively. We select the compared image downscaling models from two-stage unsupervised SISR methods in which synthesized LR images are downscaled from HR images using a trained image downscaling network or translated from Bicubic downscaled image with a trained translation network. These image downscaling models include Impressionism \cite{ji2020real}, ADL \cite{son2021toward}, DASR \cite{wei2021unsupervised} Pseudo-supervision \cite{maeda2020unpaired} and SRResCGAN \cite{umer2020deep}. Furthermore, we also compare the proposed SDFlow with state-of-the-art diverse image degradation modeling models, \ie, DeFlow \cite{wolf2021deflow} and PDM \cite{luo2022learning}. The Bicubic downscaled real-world HR images are also included as a baseline for better comparison. All of the methods are trained on the training dataset of the RealSR and DRealSR, and tested on their test dataset, respectively. Except for evaluating the downscaling results using PSNR, SSIM, LPIPS and IFC, we also employ the Fréchet inception distance (FID) \cite{fid} to compare distribution of the synthesized LR images with distribution of the real-world LR images. 10 samples are used to measure the diversity of the model outputs. No-reference metrics are not used since these image degradation modeling methods are not intend to generate perceptual pleasing LR images.

\begin{table*}[h]
\centering
\caption{Quantitative comparison for $4\times$ downscaling on the RealSR and DRealSR test datasets. $\uparrow$ indicates the higher score the better performance, while $\downarrow$ means the lower score the better performance.}\label{tab:lr_quantitative}
\resizebox{\textwidth}{!}{\begin{tabular}{c|cccccc||cccccc} 
\hline
\multirow{2}{*}{\begin{tabular}[c]{@{}c@{}}Downscaling\\methods\end{tabular}} & \multicolumn{6}{c||}{RealSR}                                                                                      & \multicolumn{6}{c}{DRealSR}                                                                                        \\ 
\cline{2-13}
                                                                              & PSNR$\uparrow$   & SSIM$\uparrow$  & LPIPS$\downarrow$ & IFC$\uparrow$   & FID$\downarrow$  & Diversity$\uparrow$ & PSNR$\uparrow$   & SSIM$\uparrow$  & LPIPS$\downarrow$ & IFC$\uparrow$   & FID$\downarrow$  & Diversity$\uparrow$  \\ 
\hhline{=======::======}
Bicubic                                                                       & 31.1163          & 0.9187          & 0.1503            & 4.5893          & 44.1883          & -                   & 32.5460          & 0.9215          & 0.1871            & 3.8584          & 34.1340          & -                    \\
Impressionism                                                                 & 31.6296          & 0.9439          & 0.0836            & 4.6893          & 30.5156          & -                   & 33.6976          & 0.9456          & 0.1126            & \textbf{4.1153} & 23.2060          & -                    \\
ADL                                                                           & 34.5064          & 0.9659          & 0.0521            & \textbf{5.2747} & 25.5383          & -                   & 33.9497          & 0.9484          & 0.0932            & 4.0052          & 17.6886          & -                    \\
DASR                                                                          & 33.7105          & 0.9591          & 0.0755            & 4.9142          & 29.7748          & -                   & 33.3525          & 0.9478          & 0.1075            & 4.0010          & 20.2454          & -                    \\
Pseudo-supervision                                                            & 32.4882          & 0.946           & 0.0566            & 3.8569          & 24.3562          & -                   & 33.7735          & 0.9468          & 0.0872            & 3.7509          & 17.0282          & -                    \\
SRResCGAN                                                                     & 31.6799          & 0.9296          & 0.0867            & 4.4058          & 35.4208          & -                   & 30.1125          & 0.8719          & 0.1739            & 2.9817          & 33.1498          & -                    \\ 
\hline
DeFlow                                                                        & 30.5338          & 0.9421          & 0.0726            & 4.2362          & 42.2095          & \textbf{10.1595}    & 30.3015          & 0.9138          & 0.1443            & 3.3259          & 37.2236          & \textbf{8.3247}      \\
PDM                                                                           & 26.9306          & 0.7974          & 0.1448            & 2.4972          & 39.4940          & 3.9635              & 29.5434          & 0.9114          & 0.1493            & 3.3518          & 25.3235          & 5.3953               \\
SDFlow                                                                        & \textbf{34.8498} & \textbf{0.9665} & \textbf{0.0439}   & 4.9370          & \textbf{22.2988} & 2.2314              & \textbf{34.6937} & \textbf{0.9533} & \textbf{0.0790}   & 4.0524          & \textbf{15.0068} & 2.2387               \\
\hline
\end{tabular}}
\end{table*}
Table \ref{tab:lr_quantitative} shows the quantitative comparison of $4\times$ downscaled HR images using various image downscaling methods on the RealSR and DRealSR test datasets. The proposed SDFlow achieves the best performance on the PSNR, SSIM, LPIPS, and FID metrics and the second-best performance on the IFC metric. The lower FID score means that the distribution of synthesized LR images is closer to the distribution of the ground truth. Thus, the LR images generated by the proposed SDFlow are more realistic as the real-world LR images than that of the compared image downscaling methods. As to the diversity metric, DeFlow and PDM obtain higher score. However, purely high diversity does not indicate high accurate diverse results \cite{diversity}. The high diversity score is attributed to the added quantization noise \cite{wolf2021deflow} synthetic noise \cite{luo2022learning}. We experimentally find that when noise is removed, diversity score of DeFlow and PDM drops significantly below that of the SDFlow. For the other compared image downscaling methods, ADL obtains the second best performance on the PSNR and SSIM metrics, while Pseudo-supervision achieves the second best FID score.

\begin{figure*}
    \centering
    \subfloat{\includegraphics[width=\textwidth]{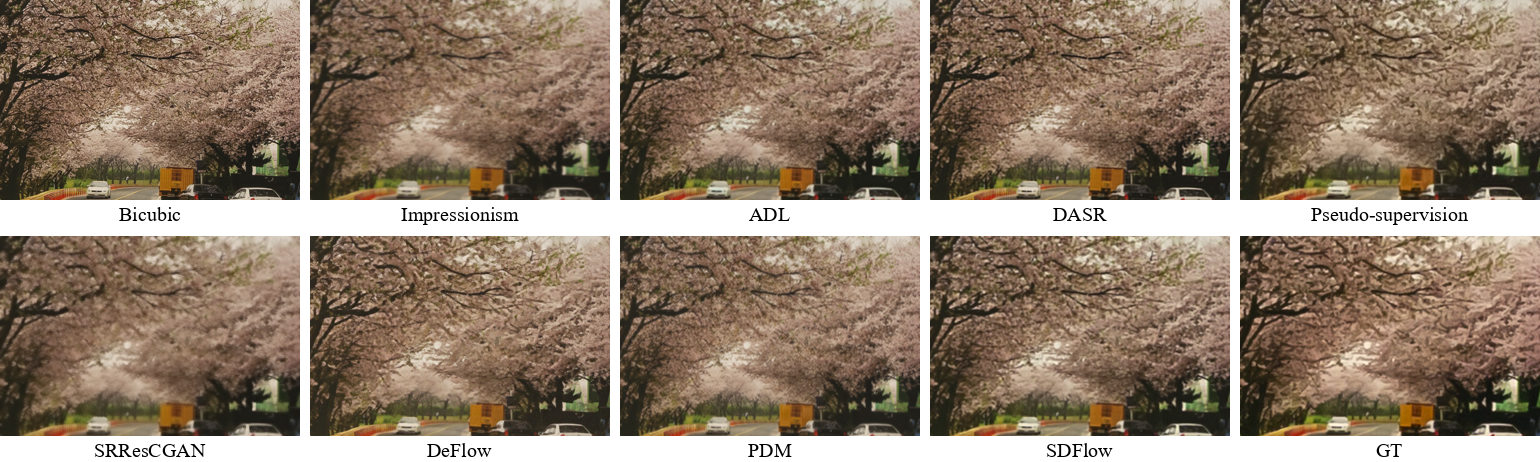}}\\
    \vspace{-9pt}
    \subfloat{\includegraphics[width=\textwidth]{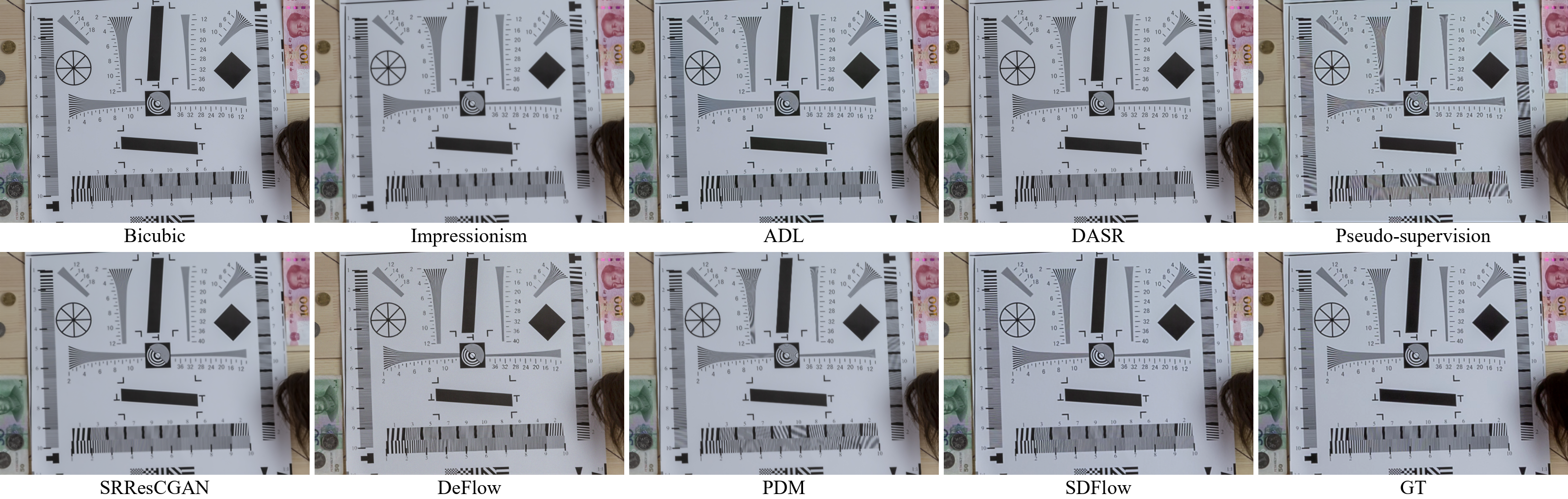}}
    \caption{Qualitative $4\times$ downscaled real-world HR images on the RealSR test dataset. Sample images are selected with typical complex textures and clear structural contents.}\label{fig:lr_qualitative}
\end{figure*}

Additionally, we also present visual comparison of downscaled real-world HR images generated by different image downscaling methods. \fig{\ref{fig:lr_qualitative}} illustrates the $4\times$ downscaled real-world HR images on the RealSR test dataset. Sample images are selected with complex textures and structural contents to better compare degradation effects. From \fig{\ref{fig:lr_qualitative}} we can observe that the Bicubic downscaled HR images are sharper than that of the other methods, since Bicubic can be treated as the ideal downscaling with a simple and limited degrees of degradation. LR images synthesized by the Impressionism and SRResCGAN model are blurry than ground-truth images, while DASR and PDM generate sharper LR images than the ground-truth. LR images generated by ADL, Pseudo-supervision, DeFlow, and the proposed SDFlow are more perceptually similar to the real-world LR images than the other compared methods.

Another way to evaluate the quality of the downscaled images against the real-world LR images is to train the same SISR model using paired real-world HR/LR images and synthesize HR/LR images, then compare their SR performance. We compare the quality of degraded LR images produced by the proposed SDFlow with two state-of-the-art unpaired diverse image degradation modeling methods, \ie, the DeFlow \cite{wolf2021deflow} and PDM \cite{luo2022learning}. We train ESRGAN \cite{esrgan} on synthesized HR/LR image pairs generated by the three models. We also train ESRGAN on the real-world HR/LR image pairs of the RealSR and DRealSR as a baseline for a better comparison. All of the trained ESRGAN models are tested on the RealSR and DRealSR test dataset. Table \ref{tab:lr2sr_quantitative} shows the quantitative $4\times$ SR results of LR images generated by different downscaling methods. As can be observed, the SR performance on LR images generated by the proposed SDFlow achieves the best on the PSNR metric, and DeFlow achieves the best on the SSIM and IFC metrics. However, PSNR, SSIM and IFC are not suitable for evaluating GAN based SR methods \cite{jinjin2020pipal}, thus, they are used only as a reference. As to the perceptual-oriented LPIPS and no-reference visual quality evaluation metrics, ESRGAN trained on real-world paired HR/LR image pairs obtains the best performance. The proposed SDFlow achieves the second-best performance on the perceptual-oriented metrics and obtains comparable performance on the full-reference metric with that of  ESRGAN trained on real-world HR/LR image pairs.
\begin{table*}[h]
\centering
\caption{Quantitative comparison for $4\times$ SR on the RealSR and DRealSR test datasets. The SR model used is the ESRGAN. $\uparrow$ indicates the higher score the better performance, while $\downarrow$ means the lower score the better performance.}
\label{tab:lr2sr_quantitative}
\resizebox{\textwidth}{!}{\begin{tabular}
{c|cccc|ccc||cccc|ccc} 
\hline
\multirow{3}{*}{\begin{tabular}[c]{@{}c@{}}Downscaling\\methods\end{tabular}} & \multicolumn{7}{c||}{RealSR}                                                                                                                                                                 & \multicolumn{7}{c}{DRealSR}                                                                                                                                                                   \\ 
\cline{2-15}
                                                                              & \multicolumn{4}{c|}{Full-reference}                                                                        & \multicolumn{3}{c||}{No-reference}                                              & \multicolumn{4}{c|}{Full-reference}                                                                        & \multicolumn{3}{c}{No-reference}                                                 \\ 
\cline{2-15}
                                                                              & PSNR$\uparrow$            & SSIM$\uparrow$           & LPIPS$\downarrow$        & IFC$\uparrow$            & NIQE$\downarrow$         & PIQE$\downarrow$          & NRQM$\uparrow$           & PSNR$\uparrow$            & SSIM$\uparrow$           & LPIPS$\downarrow$        & IFC$\uparrow$            & NIQE$\downarrow$         & PIQE$\downarrow$          & NRQM$\uparrow$            \\ 
\hhline{========::=======}
\begin{tabular}[c]{@{}c@{}}Real-world\\paired LR\end{tabular}                 & 27.2406 & 0.7628                   & \textbf{0.2132}  & 1.0632                   & 5.2841 & \textbf{44.7392}  & \textbf{5.6829}  & 29.6088 & 0.8110                   & \textbf{0.2434}  & 0.7673                   & {5.8348}  & \textbf{40.9859}  & \textbf{4.9613}   \\ 
\hline
DeFlow                                                                        & 26.9231                   & \textbf{0.7942}  & 0.3356                   & \textbf{1.1154}  & 7.2015                   & 75.4913                   & 3.3002                   & 28.7793                   & \textbf{0.8554}  & 0.3734                   & \textbf{0.8279}  & 9.1133                   & 78.2641                   & 2.7917                    \\
PDM                                                                           & 26.5699                   & 0.7614                   & 0.2634                   & 1.0041                   & 5.4842                   & 51.7233                   & 4.6282                   & 29.0866                   & 0.8457                   & 0.3332                   & 0.8072                   & 7.4917                   & 54.4887                   & 3.1469                    \\
SDFlow                                                                        & \textbf{27.3099}  & 0.7674 & 0.2262 & 1.0746 & \textbf{5.2521}  & 46.3027 & 5.5458 & \textbf{30.2033}  & 0.8483 & 0.2832 & 0.8144 & 7.0189 & 50.2428 & 3.7117  \\
\hline
\end{tabular}}
\end{table*}

\begin{figure*}
\centering
    \subfloat{\includegraphics[width=\textwidth]{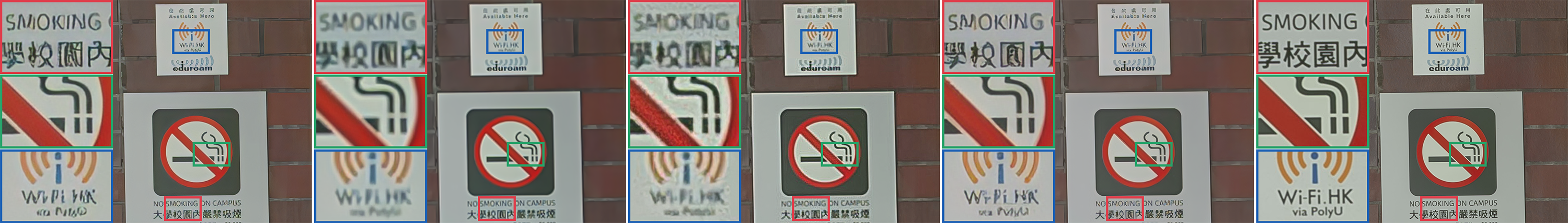}}\\
    \vspace{-9pt}
    \subfloat{\includegraphics[width=\textwidth]{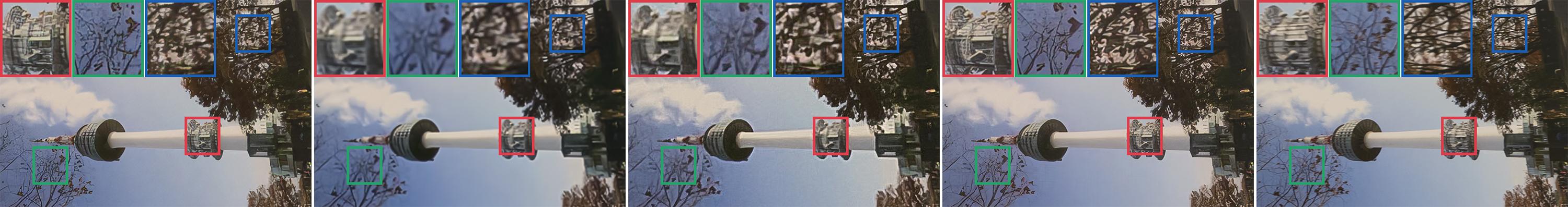}}\\
    \vspace{-9pt}
    \subfloat{\includegraphics[width=\textwidth]{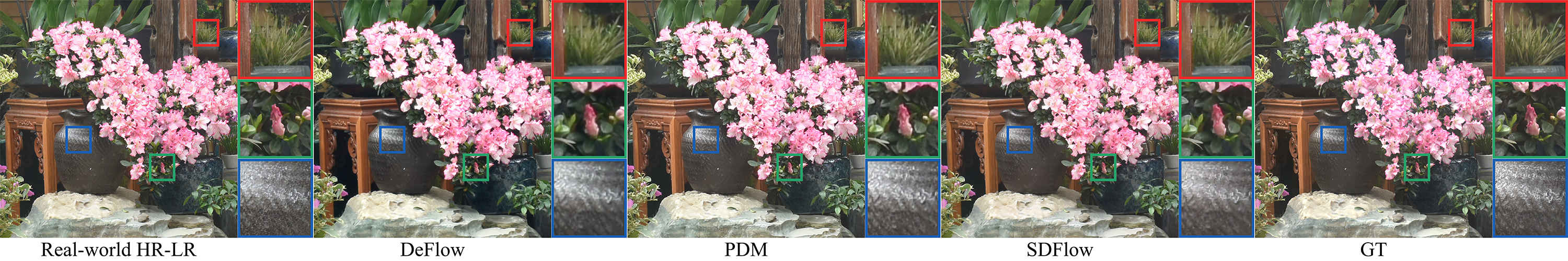}}
    \caption{Qualitative $4\times$ SR images on the RealSR and DRealSR test dataset. Three parts of each SR image are magnified for better visual comparison. The SR model used is the ESRGAN and name of each column indicates that the ESRGAN is trained with LR images generated by the named method. Visual quality of SR images generated by the ESRGAN trained on HR/LR images synthesized by the proposed SDFlow model is comparable to the ESRGAN trained on real-world HR/LR image pairs.}\label{fig:qualitative_LR2SR}
\end{figure*}

\fig{\ref{fig:qualitative_LR2SR}} illustrates the visual results of $4\times$ SR images generated by ESRGAN trained on HR/LR image pairs synthesized by each of the diverse image degradation modeling methods. Local regions that are abundant in textures or with structural content are zoomed in for better perceptual comparison. As can be observed, the visual quality of SR images generated by ESRGAN trained on HR/LR image pairs synthesized by the proposed SDFlow is comparable to that of ESRGAN trained on real-world HR/LR image pairs, which mostly resembles ground truth. As to the visual quality of SR images of ESRGAN trained on HR/LR image pairs synthesized by DeFlow and PDM, the SR results of LR images generated by PDM have sharper edges but also have more artifacts than that of DeFlow. In conclusion, the LR images produced by the proposed SDFlow are more realistic as the real-world LR images than that of the compared diverse image degradation methods in terms of SR using the same SR architecture.

\subsection{Model analysis and ablation studies}
In this section, we conduct ablation studies to investigate the effects of different model configurations on the performance of the proposed SDFlow. Ablation studies are grouped into two parts, the first part is ablation studies on the Super-resolution Flow and the second part is ablation studies on the Downscaling Flow.

\begin{table*}[h]
\centering
\caption{Ablation studies on the latent space for $4\times$ SR of the Super-resolution Flow. Results are tested on SDFlow trained with forward loss. Following \cite{liang2021hierarchical}, distortion-oriented metrics (PSNR, SSIM) are tested with $\tau=0$ and perceptual-oriented metrics are tested with $\tau=0.8$. $\uparrow$ indicates the higher score the better performance, while $\downarrow$ means the lower score the better performance.}
\label{tab:ablation_sr}
\resizebox{\textwidth}{!}{\begin{tabular}{c|ccccc|ccc||ccccc|ccc} 
\hline
\multirow{3}{*}{}                                                    & \multicolumn{8}{c||}{RealSR}                                                                                                                           & \multicolumn{8}{c}{DRealSR}                                                                                                                             \\ 
\cline{2-17}
                                                                     & \multicolumn{5}{c|}{Full-reference}                                                            & \multicolumn{3}{c||}{No-reference}                    & \multicolumn{5}{c|}{Full-reference}                                                            & \multicolumn{3}{c}{No-reference}                       \\ 
\cline{2-17}
                                                                     & PSNR$\uparrow$   & SSIM$\uparrow$  & LPIPS$\downarrow$ & IFC$\uparrow$   & Diversity$\uparrow$ & NIQE$\downarrow$ & PIQE$\downarrow$ & NRQM$\uparrow$  & PSNR$\uparrow$   & SSIM$\uparrow$  & LPIPS$\downarrow$ & IFC$\uparrow$   & Diversity$\uparrow$ & NIQE$\downarrow$ & PIQE$\downarrow$ & NRQM$\uparrow$   \\ 
\hhline{=========::========}
\begin{tabular}[c]{@{}c@{}}w/o content\\loss\end{tabular}            & 15.5997          & 0.5967          & 0.7849            & 0.2151          & -                   & -                & -                & -               & 16.0491          & 0.7327          & 0.7161            & 0.2309          & -                   & -                & -                & -                \\
\begin{tabular}[c]{@{}c@{}}w/o content\\discriminator\end{tabular}   & 20.7647          & 0.6657          & 0.5750            & 0.4874          & 8.3426              & 7.8056           & 62.5859          & 3.0516          & 26.5049          & 0.8420          & 0.4998            & 0.7588          & 3.3082              & 9.2056           & 81.2649          & 2.6908           \\
\begin{tabular}[c]{@{}c@{}}Direct\\Gaussianization\end{tabular}      & 28.3252          & 0.8054          & 0.2606            & 0.9766          & \textbf{14.2803}    & 6.1460           & 61.2127          & 4.5089          & 30.6693          & 0.8692          & 0.2903            & 0.7891          & \textbf{12.3829}    & 7.8669           & 75.0254          & 3.2559           \\
\begin{tabular}[c]{@{}c@{}}Conditional\\Gaussianization\end{tabular} & 28.4586          & 0.8099          & 0.2353            & 1.0185          & 14.0293             & 5.8799           & 57.2166          & 4.5649          & 30.7752          & 0.8709          & 0.2864            & 0.8065          & 11.3952             & 7.7682           & 73.9877          & 3.3323           \\
Full model                                                           & \textbf{28.5813} & \textbf{0.8118} & \textbf{0.2321}   & \textbf{1.0448} & 11.9467             & \textbf{5.7145}  & \textbf{55.3054} & \textbf{4.6572} & \textbf{31.0747} & \textbf{0.8726} & \textbf{0.2852}   & \textbf{0.8368} & 10.0632             & \textbf{7.6900}  & \textbf{71.2724} & \textbf{3.4047}  \\
\hline
\end{tabular}}
\end{table*}

\begin{figure*}
    \centering
    \includegraphics[width=\textwidth]{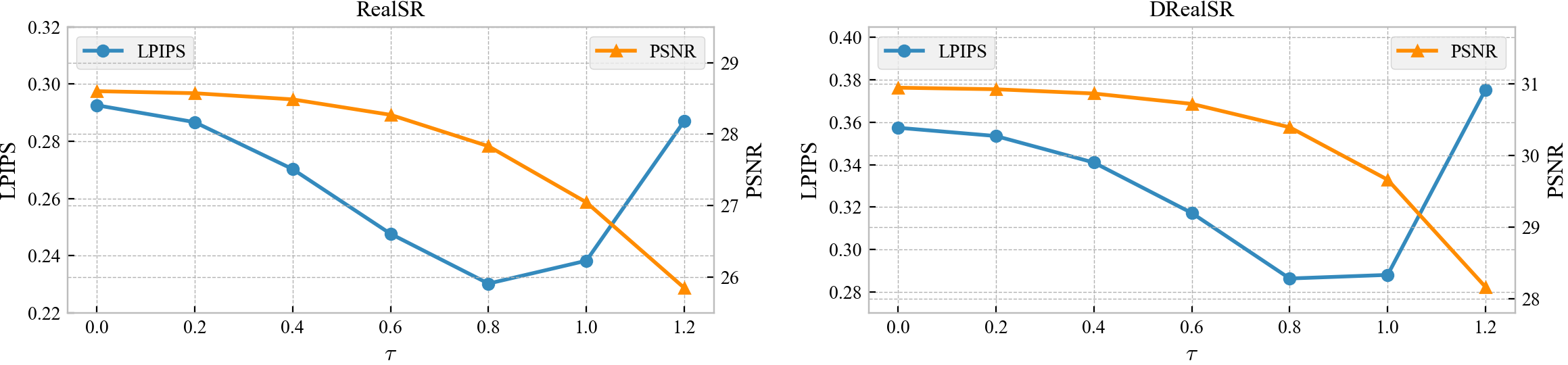}
    \caption{Correlations between the sampling temperature $\tau$ and performance of the SDFlow in terms of PSNR and LPIPS.}
    \label{fig:temp-lpips}
\end{figure*}
\subsubsection{Ablation studies on the Super-resolution Flow}
As the main idea of the proposed SDFlow is to decouple content information from the HR and LR images and map them to the same latent space. Therefore, we analyze the effects of latent space learning on the performance of the SDFlow. First, we remove the content loss (\eq{\ref{eq:lr_content}}) in training the SDFlow. As shown in \texttt{w/o content loss} in Table \ref{tab:ablation_sr}, the model did not converge since the flow network cannot determine which part is the content when factorizing the latent variable. With the employment of the content loss, the SDFlow can learn to capture the content information close to the low-pass filtered LR and BD downscaled HR image. However, the domain gap between the content latent variables of LR and HR images still exists. Thus, as shown in \texttt{w/o content discriminator} in Table \ref{tab:ablation_sr}, without the content discriminator (\eq{\ref{eq:domain}}), the model performs poorly. Next, we analyze the distribution modeling of the HF components. As discussed in Section \ref{sec:model_architecture}, we use a conditional flow branch to further transform the HF latent variable into a standard normal distribution. To validate the effects, we remove the conditional flow branch and directly Gaussianize the HF latent variable. Results are reported as \texttt{Direct Gaussianization} in Table \ref{tab:ablation_sr}, due to the lack of sufficient modeling ability, direct Gaussianization leads to inferior performance when compared to the full model. Further, we analyze the condition effects provided by the content latent variable, based on the direct Gaussianization, we add a 1 conditional flow step to transform the HF latent variable into the standard normal distribution. As shown in \texttt{Conditional Gaussianization} in Table \ref{tab:ablation_sr}, the added conditional variable improves the performance of SR.

Here, we analyze the effects of the sampling temperature $\tau$ on SR performance of the proposed SDFlow in terms of PSNR and LPIPS. The sampling temperature controls the variance of the latent variable of the HF components when generating the SR image. As shown in \fig{\ref{fig:temp-lpips}}, the decreased sampling temperature, \ie, $\tau < 1$, increases the image quality in terms of LPIPS, and the best LPIPS is obtained when $\tau=0.8$. The sampling procedure becomes deterministic when $\tau$ decreases to 0, which achieves the best performance on the PSNR metric. However, the generated SR images are blurry as illustrated in \fig{\ref{fig:qualitative_SR}}. Increasing the sampling temperature leads to notable improvements in perceptual quality in terms of the LPIPS metric while degrades the distortion oriented PSNR metric. When the sampling temperature is set to larger than 1, we can observe a significant performance drop on the LPIPS and PSNR metric. \fig{\ref{fig:temp-lpips}} also shows the perception-distortion trade-off of the SR provided by the SDFlow, and it can be controlled by adjusting the sampling temperature.

\begin{figure*}
    \centering
    \includegraphics[width=\textwidth]{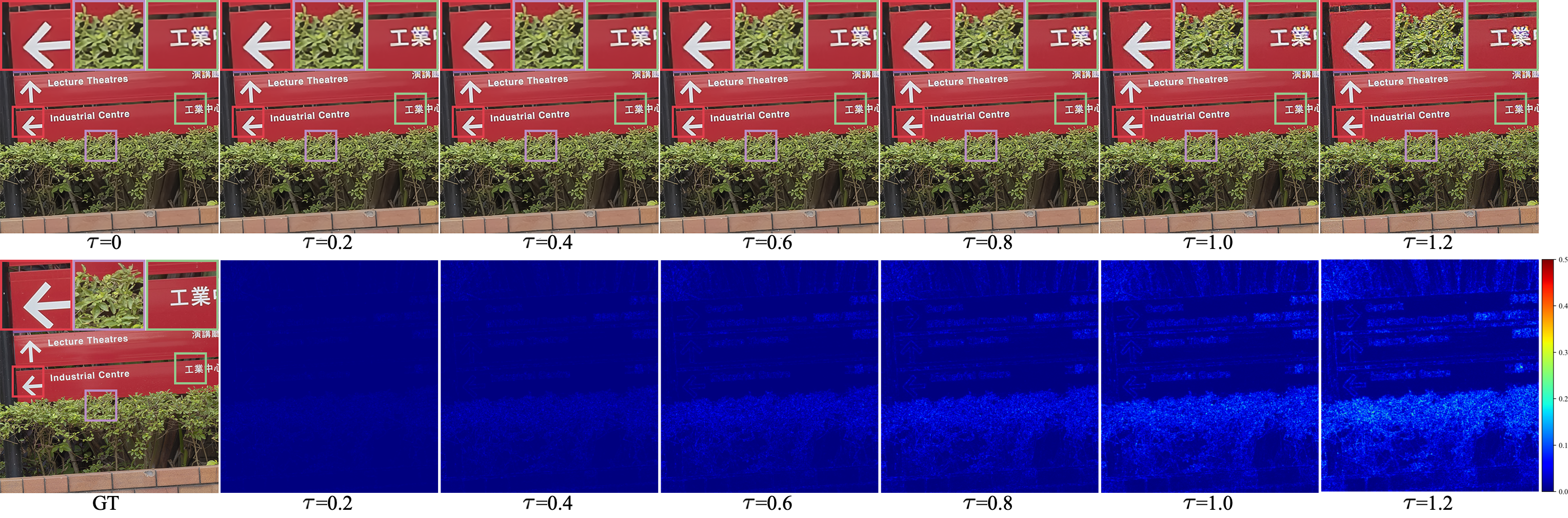}
    \caption{Visual comparison of SR images generated with different sampling temperatures, and difference between SR images generated with $\tau\neq 0$ and SR image generated with $\tau=0$. Three regions of the RGB images are magnified for better visual comparison.}\label{fig:sr_diff}
\end{figure*}

To intuitively investigate the effects of the sampling temperature on the generated SR image, we present visual comparison of SR images generated with different sampling temperature in \fig{\ref{fig:sr_diff}}. Additionally, we also calculate the difference between SR images generated with $\tau\neq 0$ and SR images generated with $\tau=0$. We can observe that the different sampling temperatures only focus on adding high-frequency contents with different strength. With the increase of the sampling temperature, more perceptible high-frequency contents are produced. However, large sampling temperature will lead to exaggerated textures and distorted edges.

\subsubsection{Ablation studies on the Downscaling Flow}
Here we conduct ablation studies on the Downscaling Flow of the SDFlow. Similar to the ablation studies for the Super-resolution Flow, we first remove the content loss (\eq{\ref{eq:lr_content}}) from the training objectives. As shown in \texttt{w/o content loss} in Table \ref{tab:ablation_lr}, the model failed to converge since the content feature extractor cannot learn to keep content information without any guidance. Then, we remove the content domain discriminator to validate the effects of the remaining content domain gap on the performance of the downscaling model. As shown in \texttt{w/o content discriminator} in Table \ref{tab:ablation_lr}, performance of the Downscaling Flow drops significantly. LR images are produced by the Downscaling Flow, which takes content latent variable of the HR image and sampled degradation information as input. To validate the effects of the sampled degradation information, we generate LR images without adding the sampled degradation. As shown in \texttt{w/o degradations} in Table \ref{tab:ablation_lr}, generating downscaled images without sampled degradations undermines the performance of the Downscaling Flow, especially on the FID metric. Finally, we investigate the effectiveness of the proposed base distribution modeling of the transformed degradation information using learned mixture of Gaussians. As shown in \texttt{Standard Normal Distribution} in Table \ref{tab:ablation_lr}, modeling the base distribution of the degrdation information using the standard normal distribution obtains slightly inferior performance on the full-reference metrics. Modeling the base distribution of the degradation information using the mixture of Gaussians improves the FID score.
\begin{table*}[h]
\centering
\caption{Ablation studies on the latent space for $4\times$ SR of the Super-resolution Flow. Results are tested on SDFlow trained with forward loss. Following \cite{liang2021hierarchical}, distortion-oriented metrics (PSNR, SSIM) are tested with $\tau=0$ and perceptual-oriented metrics are tested with $\tau=0.8$. $\uparrow$ indicates the higher score the better performance, while $\downarrow$ means the lower score the better performance.}
\label{tab:ablation_lr}
\resizebox{\textwidth}{!}{\begin{tabular}{c|cccccc||cccccc} 
\hline
\multirow{2}{*}{}            & \multicolumn{6}{c||}{RealSR}                                                                                      & \multicolumn{6}{c}{DRealSR}                                                                                        \\ 
\cline{2-13}
                             & PSNR$\uparrow$   & SSIM$\uparrow$  & LPIPS$\downarrow$ & IFC$\uparrow$   & FID$\downarrow$  & Diversity$\uparrow$ & PSNR$\uparrow$   & SSIM$\uparrow$  & LPIPS$\downarrow$ & IFC$\uparrow$   & FID$\downarrow$  & Diversity$\uparrow$  \\ 
\hhline{=======::======}
w/o content loss             & 14.2424          & 0.2606          & 0.6935            & 0.1947          & -                & -                   & 15.1371          & 0.4414          & 0.6994            & 0.2603          & -                & -                    \\
w/o degradations             & 32.5555          & 0.9447          & 0.1130            & 4.7147          & 45.6483          & -                   & 33.6884          & 0.9424          & 0.1123            & 3.8736          & 38.1845          & -                    \\
w/o content discriminator    & 17.4049          & 0.4785          & 0.4575            & 2.6089          & 83.5019          & \textbf{4.1993}     & 23.3012          & 0.6376          & 0.4944            & 1.9687          & 71.2011          & \textbf{6.6017}      \\
Standard Normal Distribution & 34.6329          & 0.9648          & 0.0449            & 4.9109          & 23.8114          & 2.7137              & 34.4680          & 0.9513          & 0.0821            & 4.0311          & 16.3796          & 2.6696               \\
Full model                   & \textbf{34.7852} & \textbf{0.9665} & \textbf{0.0444}   & \textbf{4.9730} & \textbf{22.6374} & 2.7934              & \textbf{34.5026} & \textbf{0.9521} & \textbf{0.0824}   & \textbf{4.0338} & \textbf{15.7554} & 2.7269               \\
\hline
\end{tabular}}
\end{table*}

\begin{figure*}
    \centering
    \includegraphics[width=\textwidth]{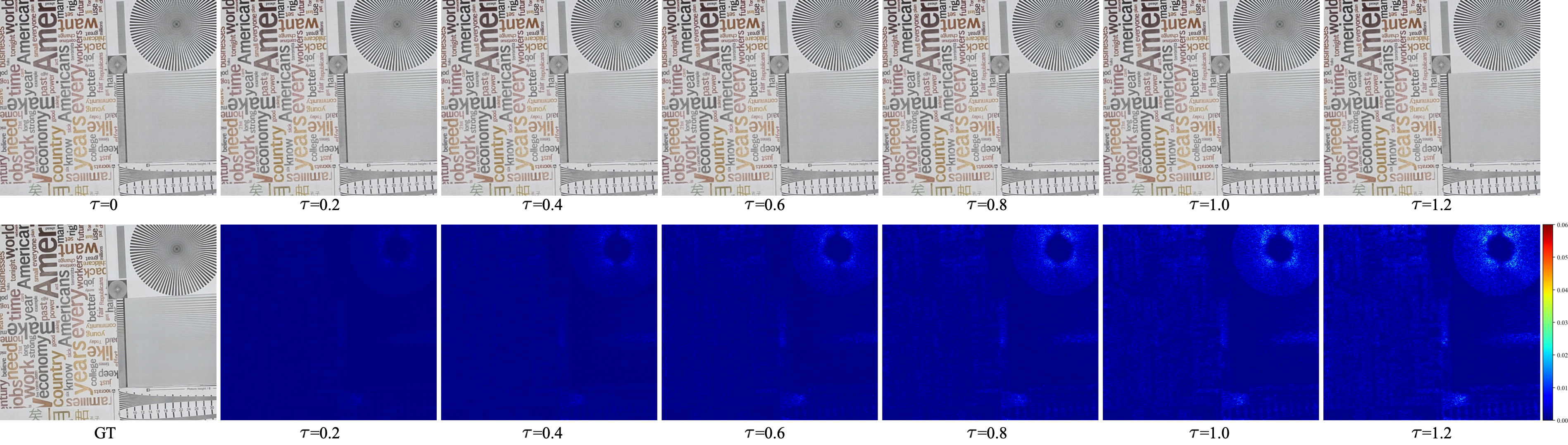}
    \caption{Visual comparison of downscaled images generated with different sampling temperatures, and difference between downscaled images generated with $\tau\neq 0$ and downscaled image generated with $\tau=0$.}\label{fig:ds_diff}
\end{figure*}

One of the prominent features of the SDFlow is that it learns the conditional distribution of the degradation information given the image content latent variable, and thereby it can generate diverse downscaled (degraded) LR images of the input HR image. Also, the Downscaling Flow of the SDFlow can generate LR images with degradations at different strengths, which is achieved by sampling the standard normal distribution of the transformed degradation latent variables with different sampling temperatures. \fig{\ref{fig:ds_diff}} illustrates the visual results of the downscaled images generated with different sampling temperatures. Differences between downscaled images generated with different sampling temperatures are barely visual perceptible. Thus, we also calculate the difference between downscaled images generated with $\tau\neq 0$ and downscaled image generated with $\tau=0$. As shown in \fig{\ref{fig:ds_diff}}, differences between downscaled images generated with different sampling temperatures are mainly located in regions with high-frequency contents. Additionally, with the increasing of    sampling temperature, the degradation effects are enhanced. When $\tau<1$, nuanced degradations are imposed, and $\tau>=1$  amplified the degradations.

\section{Conclusion}\label{sec:conclusion}
Image super-resolution and downscaling modelling in real-world scenarios is challenging due to the ill-posed nature of the task and lack of ideal paired training data. In this paper, we propose an invertible model, named SDFlow, which simultaneously learns the bidirectional many-to-many mapping for real-world image super-resolution and downscaling from unpaired data. The main idea of the proposed method is to discover the shared image content representation space between real-world HR and LR images, and to model the distribution of the remaining HF information and degradation information in the latent space. We derive the solution under the variational inference framework and solve it using the invertible SDFlow, in which the Super-resolution Flow decouples the HR image into content and HF latent variables, and the Downscaling Flow decouples the LR image into content and degradation latent variables. Content representations of the HR and LR image are mapped into the shared latent space under the GAN framework, HF and degradation latent variables are transformed into an easy-to-sample distribution space using the conditional normalizing flow. Diverse image super-resolution and downscaling are realized by first obtaining the content representation using the forward pass of the Super-resolution Flow or the Downscaling Flow; then latent representation of the HF and degradation are sampled conditioned on the content representation; SR image and the downscaled image are generated using the backward pass of the Super-resolution Flow and Downscaling Flow by taking the combination of content and HF or degradation latent variables as input, respectively. Experimental results indicate that the proposed SDFlow model achieves the state-of-the-art SR performance in unpaired learning settings, while it can still generate more diverse realistic real-world LR images.

\normalsize
\bibliography{ref}


 




\vfill

\end{document}